\documentclass[letterpaper]{article}

\usepackage{arxiv}

\usepackage{amsmath}
\usepackage{amsfonts}
\usepackage{amssymb}

\usepackage{algorithm}
\usepackage{algpseudocode}

\usepackage[utf8]{inputenc} 
\usepackage[T1]{fontenc}    
\usepackage[hidelinks]{hyperref}
\usepackage[nameinlink,capitalize]{cleveref}
\usepackage{url}            
\usepackage{booktabs}       

\usepackage{nicefrac}       
\usepackage{microtype}      
\usepackage{graphicx}
\usepackage{caption}
\usepackage[numbers]{natbib}
\usepackage{doi}

\usepackage[section]{placeins} 

\usepackage[dvipsnames]{xcolor}
\usepackage{makecell}
\usepackage[version=4]{mhchem}
\usepackage[per-mode=symbol,round-precision=3,scientific-notation=false,range-phrase = \text{--}]{siunitx}

\usepackage{wrapfig}
\usepackage{adjustbox}

\usepackage{framed} 
\usepackage{multicol} 
\usepackage{nomencl} 
\usepackage{ifthen}
\renewcommand\nomgroup[1]{%
  \ifthenelse{\equal{#1}{A}}{%
    \item[\textbf{Acronyms}]}{
  \ifthenelse{\equal{#1}{R}}{
    \item[\textbf{Roman Symbols}]}{
  \ifthenelse{\equal{#1}{G}}{%
    \item[\textbf{Greek Symbols}]}{
  \ifthenelse{\equal{#1}{S}}{%
    \item[\textbf{Superscripts}]}{
  \ifthenelse{\equal{#1}{U}}{%
    \item[\textbf{Subscripts}]}{
  \ifthenelse{\equal{#1}{X}}{%
    \item[\textbf{Other Symbols}]}{
  {}}}}}}}}
\renewcommand*{\nompreamble}{\markboth{\nomname}{\nomname}}

\makenomenclature
\renewcommand*\nompreamble{\begin{multicols}{2}}
\renewcommand*\nompostamble{\end{multicols}}

\newcommand\T{\rule{0pt}{2.6ex}}
\newcommand\B{\rule[-1.2ex]{0pt}{0pt}}

\usepackage{xspace}
\newcommand*{\eg}{e.g.,\@\xspace}
\newcommand*{\ie}{i.e.,\@\xspace}

\newcommand*{\vs}{vs.\@\xspace}

\makeatletter
\newcommand*{\etc}{%
    \@ifnextchar{.}%
        {etc}%
        {etc.\@\xspace}%
}
\makeatother

\newcommand{\myCOMMENT}[1]{}

\DeclareUnicodeCharacter{025B}{\ensuremath{\varepsilon}}

\graphicspath{{figures/}}

\usepackage{listings}

\definecolor{bkgd}{RGB}{240,242,246}
\definecolor{ceruleanblue}{rgb}{0.16, 0.32, 0.75}
\definecolor{orange-red}{rgb}{1.0, 0.27, 0.0}
\definecolor{anotherblue}{RGB}{37,92,243}
\definecolor{blackblue}{RGB}{46,60,85}
\definecolor{goldyellow}{RGB}{199,146,12}
\lstdefinestyle{altstyle2}{
    backgroundcolor=\color{bkgd},
    basicstyle=\ttfamily\footnotesize\color{blackblue},
    breakatwhitespace=false,
    breaklines=true,
    captionpos=b,
    commentstyle=\color{goldyellow},
    keepspaces=true,
    keywordstyle=\color{orange-red},
    language=Python,
    numbersep=5pt,
    numberstyle=\tiny\color{ceruleanblue},
    showspaces=false,
    showstringspaces=false,
    showtabs=false,
    stringstyle=\color{anotherblue},
    tabsize=2,
    numbers=left
}
\newcommand{\pkg}[1]{\texttt{#1}}
\newcommand{\ith}[1]{$#1$\textsuperscript{th} }
\renewcommand{\O}[1]{$\mathcal{O}(#1)$}

\usepackage[scaled=0.86]{helvet}

\crefname{lstlisting}{listing}{listings}
\Crefname{lstlisting}{Listing}{Listings}

\crefname{algorithm}{Alg.}{Algs.}
\Crefname{algorithm}{Alg.}{Algs.}

\usepackage{pdflscape}

\DeclareMathOperator*{\argmin}{\arg\!\min}
\algnewcommand{\Inputs}[1]{%
  \State \textbf{Inputs:}
  \Statex \hspace*{\algorithmicindent}\parbox[t]{.8\linewidth}{\raggedright #1}
}
\algnewcommand{\Initialize}[1]{%
  \State \textbf{Initialize:}
  \Statex \hspace*{\algorithmicindent}\parbox[t]{.8\linewidth}{\raggedright #1}
}

\definecolor{darkorange}{HTML}{FF8C00}
\newcommand{\cf}[1]{%
  \ifnum#1>100%
    \textcolor{darkorange}{#1}%
  \else%
    \ifnum#1>75%
      \textcolor{black}{#1}%
    \else%
      \textcolor{teal}{#1}%
    \fi%
  \fi%
}

\definecolor{textred}{RGB}{170,0,0}

\usepackage{eqparbox}

\usepackage{tikz}
\usetikzlibrary{shapes.geometric, arrows}
\usepackage{pgfplots}
\pgfplotsset{compat=1.18}

\title{Physics-informed State-space Neural Networks for Transport Phenomena}

\author{ 
	Akshay J.~Dave \\
	Nuclear Science and Engineering\\
	Argonne National Laboratory\\
	Lemont, IL 60439\\
	\texttt{ajd@anl.gov} \\
	\And
	Richard B.~Vilim \\
	Nuclear Science and Engineering\\
	Argonne National Laboratory\\
	Lemont, IL 60439\\
}

\date{}

\renewcommand{\headeright}{ }
\renewcommand{\undertitle}{ }
\renewcommand{\shorttitle}{Physics-informed state-space neural networks for transport phenomena}

\begin{document}
\maketitle

\begin{abstract}

This work introduces Physics-informed State-space neural network Models (PSMs), a novel solution to achieving real-time optimization, flexibility, and fault tolerance in autonomous systems, particularly in transport-dominated systems such as chemical, biomedical, and power plants.
Traditional data-driven methods fall short due to a lack of physical constraints like mass conservation; PSMs address this issue by training deep neural networks with sensor data and physics-informing using components' Partial Differential Equations (PDEs), resulting in a physics-constrained, end-to-end differentiable forward dynamics model.
Through two in silico experiments — a heated channel and a cooling system loop — we demonstrate that PSMs offer a more accurate approach than a purely data-driven model.
In the former experiment, PSMs demonstrated significantly lower average root-mean-square errors across test datasets compared to a purely data-driven neural network, with reductions of 44 \%, 48 \%, and 94 \% in predicting pressure, velocity, and temperature, respectively.

\vspace{1.0em}

Beyond accuracy, PSMs demonstrate a compelling multitask capability, making them highly versatile.
In this work, we showcase two: supervisory control of a nonlinear system through a sequentially updated state-space representation and the proposal of a diagnostic algorithm using residuals from each of the PDEs.
The former demonstrates PSMs' ability to handle constant and time-dependent constraints, while the latter illustrates their value in system diagnostics and fault detection.
We further posit that PSMs could serve as a foundation for Digital Twins, constantly updated digital representations of physical systems. 

\vspace{1.0em}

\end{abstract}



\section{Introduction}

Autonomous systems are envisioned to achieve real-time optimization, be flexible in operation, and be fault-tolerant \cite{gamer2020autonomous}.
These capabilities necessitate a model-based approach. 
However, purely data-driven methods, which have proliferated in the literature, lack physical constraints like mass conservation, precluding their adoption by industry.
In this work, we propose a method that \textit{fuses} physical measurements from sensors with physical knowledge of the conservation laws. 
The application of this method is specifically developed for systems dominated by transport phenomena.

Transport phenomena, which describe the movement of mass, momentum, and energy, are fundamental to many engineering fields as they offer a predictive understanding of systems' behavior.
In chemical engineering, these phenomena facilitate the design and optimization of processes in reactors, separators, and heat exchangers, which are essential for chemical, fuel, and pharmaceutical production \citep{belfiore2003transport}.
Biomedical engineering applies transport phenomena to comprehend nutrient, oxygen, and waste transportation in biological systems like the circulatory system, driving the development of new medical technologies such as artificial organs \citep{fournier2011basic}.
Power plant engineering utilizes transport phenomena in optimizing heat and mass transfer in boilers and heat exchangers for efficient and safe power generation and in emission control \citep{bird2006transport}.
They are also vital in nuclear reactor design, where effective heat transfer from the reactor core to the coolant is paramount \citep{todreas2021nuclear}.
Consequently, modeling transport phenomena is crucial for the design, optimization, and enhancement of engineering systems and processes.

A set of transport equations models transport phenomena.
The solution of transport equations involves formulating functions of space and time for multiple fields (\eg pressure, velocity, temperature), given a set of boundary and initial conditions.
Classically, there have been two major approaches to solving transport equations: analytical and numerical.
The former applies to situations where the problem setting or transport phenomena are simple enough to directly use mathematical methods (\eg separation of variables) -- this is generally not the case for systems of interest described above.
The latter is used for systems with complex geometries, tightly coupled fields, time-dependent boundary/initial conditions, or parameters that vary with the field itself.
Numerical methods include finite difference, finite volume, or finite element \citep{mattiussi1997analysis}.
These methods discretize the domain into a grid, and values at fixed locations represent the unknown fields.
The transport equations are then approximated using numerical algorithms (\eg the forward Euler method, the Crank-Nicolson method) and solved iteratively until convergence is achieved. 

The numerical methods are the current state-of-the-art approach to modeling transport phenomena for engineering systems.
Their main advantage is versatility in addressing a range of complex problems with time-dependent boundary/initial conditions.
However, they have several disadvantages.
\textit{Discretization error}: numerical methods discretize the domain and approximate the transport equations.
This process introduces error, which can be reduced with finer grids -- unfortunately increasing computational cost.
\textit{Convergence}: because numerical methods involve iterative algorithms, convergence to a solution is not guaranteed and can result in numerical instability.
\textit{Static assumptions}: generally, several parameters in a transport equation must be predefined (\eg the friction factor).
These assumptions form a rigidity around the application of numerical methods: they do not provide a straightforward path to adapt to changing system conditions.
The first two disadvantages preclude numerical methods as a basis for the control of engineering systems.
A majority of model-based control algorithms assume a reduced order or surrogate model for control \citep{richalet1993industrial,hou2013model}.

In this paper, we introduce Physics-informed State-space Neural Networks for transport phenomena.
Physics-Informed Neural Networks (PINNs) are a type of artificial neural network (ANN) used to solve partial differential equations (PDEs) \citep{raissi2019physics}.
PINNs combine the expressiveness and flexibility of ANNs with the prior knowledge of physical laws and constraints encoded in PDEs.
In a PINN approach, the neural network is trained to approximate the solution of  PDEs, given appropriate boundary and initial conditions, by minimizing the difference between the network outputs and the PDE constraints.
Thus, PINNs could be a candidate for the solution of a system of PDEs representing transport phenomena.
PINNs have several advantages.
\textit{Physical priors}: PINNs incorporate prior knowledge about the system through physical constraints.
In the context of transport phenomena, prior knowledge is the conservation of mass, momentum, and energy in the spatiotemporal domain.
\textit{Mesh-free}: PINNs do not require a mesh or a grid to discretize the spatial domain, as numerical methods do. 
Instead, PINNs use a neural network to represent the solution in a continuous fashion, without the need for discretization.
First, this aspect eliminates the need for mesh generation, which can be time-consuming and challenging.
Second, it eliminates discretization error (described in the previous paragraph) arising from the mesh.
\textit{Flexible training data}: PINNs can be trained with noisy or scarce data.
Thus, sensor measurements, whose outputs have aleatoric sources of noise, may be used to train PINNs.
Additionally, the scarcity of data is an essential aspect as the number of sensor locations (\ie cost) required to utilize PINNs is low.

Specifically, we introduce physics-informed \textit{state-space} neural networks for transport phenomena.
A plurality of previous PINN work focuses on solutions of one or more PDEs, with fixed boundary and initial conditions \cite{cuomo_scientific_2022},
\begin{equation}
   x\left(z, t\right)=\mathcal{F}\left(z, t\right)~\label{eq:dyn_no_v},
\end{equation}
where $\mathcal{F}$ is a PINN, $z$ is a position vector, $t$ is time, and $x$ is the field variable(s). 
However, in a control or state estimation setting, a form representing,
\begin{equation}
   x\left(z, t, v\right)=\mathcal{F}\left(z, t, v\right)~, \label{eq:dyn_v}
\end{equation}
is needed, where $v$ represents a vector of arbitrary control actions.
The control actions modify the boundary conditions (BCs) of the problem.
In this setting, the PINN must be able to generalize to a much larger input space.
\citet{arnold2021state} attempted a similar task for a system governed by a single PDE but did not train a monolithic PINN for varying $v$ -- rather multiple PINNs corresponding to a grid of anticipated control actions.

\subsection{Objectives \& Contributions}

The objective of this work is to develop and demonstrate a method that generates a state-space model by fusing physical measurements from sensors with knowledge of the fundamental physics of the system.
We refer to the outcome as a Physics-informed State-space neural network Model (PSM), detailed in \cref{subsec:psm_form}.
With respect to the existing literature, our work has the following major contributions:
\begin{itemize}
    \item Integration of PINNs with State-Space Models: Our work introduces a method that combines PINNs with state-space modeling, using a single neural network that can accommodate multiple and continuous control actions, and output multiple fields over the entire spatial domain of the system.
    This creates a robust modeling approach that enhances accuracy and integrates physical laws with sensor data.
    This is particularly significant in systems where the underlying physical law(s) is known yet cannot be captured in a purely data-driven approach due to a lack of instrumentation (or preclusion due to extreme conditions).

    \item Application to Transport-Dominated Systems: The PSM is specifically developed for engineered systems dominated by transport phenomena, such as those found in chemical, biomedical, and power plants. 
    In these systems, there are several sensors distributed throughout the plant.
    We simultaneously apply three PDEs (the current state-of-the-art in modeling fluid transport) to constrain the dynamics of the neural network.

    \item Demonstration through Experiments: We validate the efficacy of PSMs through two in silico experiments -- a heated channel and a cooling system loop. 
    These experiments demonstrate the capability in handling complex real-world scenarios in transport-dominated systems, providing evidence of the practical applicability of our approach. 
    Our work demonstrates the robustness in handling both homoscedastic and heteroscedastic noise in sensor measurements, a common challenge in real-world data collection.
    
    \item Demonstration of Multitask Capability: The PSM's end-to-end differentiability allows them to be used for model-based control and diagnostics schemes.
    For the former, we showcase the Reference Governor supervisory control scheme and use it to constrain excessive temperatures during operational transients. 
    For the latter, the ability to compute residuals from the PDEs enables the identification of system degradations or faults.
\end{itemize}

The paper is organized as follows.
In \cref{sec:methods}, the definition and solution of the transport equations, derivation of our proposed method, and its training procedure are presented.
Two separate experimental settings were explored.
In \cref{sec:res_ch}, results for modeling a heated channel geometry are presented, and a supervisory control method is demonstrated.
In \cref{sec:res_loop}, results for modeling a cooling system are presented, and a demonstration of a capability to use our proposed method for physics-based diagnostics during a system fault.
Finally, \cref{sec:disc} contains a discussion and conclusions on advantages, disadvantages, and additional capabilities to be explored.

\section{Methods} \label{sec:methods}

\subsection{Modeling fluid transport}

There is a large corpus of equations that model fluid transport.
This work focuses on one-dimensional single-phase flow at low-speed, \ie conditions ubiquitous in engineering systems described in the introduction.
Under these conditions, the conservation of mass, momentum, and energy are modeled by the following set of PDEs \citep{hu2015advanced}: 
\begin{align}
   \frac{\partial\rho}{\partial t} &+ \frac{\partial\left(\rho u\right)}{\partial z} = 0~, \label{eq:mass}\\
   \rho\frac{\partial u}{\partial t} &+ \rho u \frac{\partial u}{\partial z} = -\frac{\partial p}{\partial z} + \rho g - \frac{f}{D_h}\frac{\rho u | u |}{2}~, \label{eq:mom}\\
   \rho C_p \frac{\partial T}{\partial t} &+ \rho C_p u \frac{\partial T}{\partial z} = q''' ~. \label{eq:nrg}
\end{align}
In all equations, $z$ is the spatial dimension, and $t$ is the temporal dimension. 
This work uses the field variable set $(p, u, T)$, where $p$ is the pressure, $u$ is the one-dimensional velocity, and $T$ is the temperature. 
Additionally, there are several parameters.
These parameters are either constants or modeled via closure relationships.
A closure relationship models density, $\rho=\rho(p,T)$, pre-defined via experiments.
Specific heat capacity is modeled by closure relationships, $C_p=C_p(p,T)$, pre-defined via experiments.
The friction factor, $f$, can be obtained via closure relationships (Moody chart) or determined via experiments.
The gravitational coefficient, $g$, is a constant that will depend on the orientation of the component(s) with the gravitational field (and its magnitude).
The hydraulic diameter, $D_h$, can be obtained via manufacturer specifications or component measurement.
The volumetric heat source, $q'''$, depends on the configuration of the thermal energy source, \eg heater power magnitude and its location.

\subsection{Numerical solution}

The approach to obtaining numerical solutions to the transport equations is discussed. 
A solver, System Analysis Module (SAM) \citep{Hu2017}, is utilized.
The SAM code specializes in the high-fidelity simulation of advanced Nuclear Power Plants.
As part of its capabilities, it can solve the transport equations, \cref{eq:mass,eq:mom,eq:nrg}.
The Finite Element Method (FEM) is utilized to solve the PDEs.
FEM works by approximating a continuous solution over a region of interest as a combination of simple functions (\eg polynomials) that are defined over a finite set of points.
The solution is then represented as a set of coefficients describing the simple functions' behavior over the entire region.
SAM uses the continuous Galerkin FEM and Lagrange shape functions.
This work uses first-order elements using piece-wise linear shape functions with the trapezoidal rule for numerical integration.
The Jacobian-Free Newton Krylov method is used to solve the system of equations.
The second-order backward difference formula is used for time integration.
Additional details on the numerical scheme are provided in \citep[\S7]{Hu2017}.

\subsection{PINN solution}

PINNs are a type of ANN that can also solve PDEs.
The modern PINN concept was initially introduced by \citet{raissi2019physics} and has since received significant adoption by the scientific machine-learning community -- reviewed exhaustively in \citep{cuomo_scientific_2022}.
PINNs solve parameterized and nonlinear PDEs in the form,
\begin{equation}
    \frac{\partial x}{\partial t} + \mathcal{N}(x;\lambda) = 0~, \label{eq:pde_gen}
\end{equation}
where $x=x(z,t)$ is the solution, and $\mathcal{N}$ is a nonlinear operator parameterized by $\lambda$.
Each transport equation, \cref{eq:mass,eq:mom,eq:nrg} can be written in this form.
A neural network is formulated to approximate the solution,
\begin{equation}
    x(z,t)\approx \mathcal{F}(z,t;\theta),~z\in\Omega,~t\in\tau~, \label{eq:nn_def}
\end{equation}
where  $\mathcal{F}$ is a neural network that is parameterized by $\theta$.
The solution is box constrained in an n-dimensional space domain, $\Omega$, and in a one-dimensional time domain, $\tau$.
The space domain specification would depend on the geometry of the transport path and whether multiple dimensions are needed.
The specification of the time domain depends on the timescale at which the system reaches a steady state following changes in BCs. 
For instance, if the system settles slowly, a large $\tau$ is required, whereas a small $\tau$ is needed if the system reaches a steady state quickly.

In a purely data-driven approach, a series of measurements would be collected and stored in sets,
\begin{align}
    \mathcal{X}_m &= \{(z_i, t_i)_{i=1}^{N}\}~, \label{eq:ds1}\\
    \mathcal{Y} &= \{(x_i)_{i=1}^{N}\}~, \label{eq:ds2} 
\end{align}
where $z_i$ is the position, $t_i$ is the time, and $x_i$ is the field value for the \ith{i} data point.
The set $\mathcal{X}_m$ contains positions and times, where subscript $m$ denotes measurement data.
The set $\mathcal{Y}$ contains measured field values.
The superscript $N$ is the dataset size.
In a supervised learning setting, $\mathcal{X}_m$ is the input dataset, and $\mathcal{Y}$ is the output dataset.
Formally, $x_i\in\mathbb{R}^m$, where $m$ is the number of fields, $z_i\in\mathbb{R}^n\cap\Omega$ where $n$ is the number of spatial dimensions, and $t_i\in\mathbb{R}\cap\tau$ is a scalar.
A corresponding loss function is defined,
\begin{equation}
    \mathcal{L}_m = f_\mathrm{loss}\left(\mathcal{F}(\mathcal{X}_m),\mathcal{Y}\right), \label{eq:meas_loss}
\end{equation}
where $f_\mathrm{loss}$ is a loss function (\eg the mean squared error), and the $\mathcal{L}_m$ denotes the measurement loss.
The first argument of the loss function is the predicted value, and the second argument is the target value.
Thus, $\mathcal{L}_m$ quantifies the error (or loss) between predictions from the neural network and the measured field values.
The loss is then backpropagated \citep{rojas1996backpropagation} to the neural network's parameters, $\theta$.
In the purely data-driven setting, the values of $\theta$ are strictly determined by the measured field values.

In a PINN setting, an additional source of information is added.
The dynamics of the system being modeled are approximately known.
For example, if a liquid is advected in a network of pipes, the transport equations can inform the neural network that, \eg energy conservation must hold.
Thus, the transport PDEs inform the neural network if there are inconsistencies in its solution.
How is this achieved?
First, the  exact solution in \cref{eq:pde_gen} is replaced by that approximated by the neural network in \cref{eq:nn_def},
\begin{equation}
    \frac{\partial \mathcal{F}}{\partial t} + \mathcal{N}\left(\mathcal{F};\lambda\right)=0~.\label{eq:pinn_gen}
\end{equation}
To evaluate the PDEs, \eg \cref{eq:mass,eq:mom,eq:nrg}, the gradients of $\partial_t u, \partial_z u,$ \etc are needed.
In the numerical approach, the gradients are obtained by discretizing the space-time domain and approximating them using finite difference.
The PINN approach uses a mesh-free method to evaluate the gradients.
This is achieved through \textit{forward mode} automatic differentiation, where the derivative of any output of $\mathcal{F}$ is obtainable with respect to any input.
The physics-informed loss function is defined as,
\begin{equation}
    \mathcal{L}_p = f_\mathrm{loss}\left(\frac{\partial \mathcal{F}}{\partial t}(\mathcal{X}_p) + \mathcal{N}\left(\mathcal{F}(\mathcal{X}_p), \lambda\right), \emptyset \right)~, \label{eq:pinn_loss}
\end{equation}
where $\emptyset$ is a zero set. 
The target output is always zero to enforce that the neural network satisfies the right-hand side of \cref{eq:pinn_gen}.
It is important to emphasize that the target output is a zero set rather than measured field values.
Thus, $\mathcal{L}_p$ can be evaluated at arbitrary inputs sampled from the space-time domain, $(z_p, t_p) \sim \Omega\times\tau$.
In this way, the known physics of the system regularizes the neural network to the space-time domain outside of the measurement set, $\mathcal{X}_m$.
The additional set of inputs is referred to as collocation points,
\begin{equation}
    \mathcal{X}_p = \{(z_p, t_p)_{i=1}^{N}\}~. \label{eq:ds_pinn}\\
\end{equation}
Now, during training, the parameters of the neural network, $\theta$, can be updated to minimize both $\mathcal{L}_m$ and $\mathcal{L}_p$.
Thus, knowledge gathered from measurements of the system is complemented with a priori knowledge of the physics of the system.
The resulting neural network is defined as a PINN.
Measurements of a facility require the installation of sensors (\eg thermocouples and pressure transducers) and associated operation and maintenance costs.
The amount of sensors and their locations is itself an optimization task \cite{joshi2008sensor,khalid2021real}.
The PINN approach is particularly powerful when system measurements are sparse or physically precluded \cite{raissi2017physics} (\eg extreme temperature and pressure environments).

\subsection{State-space formulation (PSM)}\label{subsec:psm_form}

The objective of this work is to apply aspects of PINN to obtain a nonlinear dynamical model as an alternative to traditional linear state-space models.
A discrete-time state-space model is represented by,
\begin{align}
    x_{k+1} &= f\left(x_k, v_k\right)~,\label{eq:gen_dynamics}\\
    y_{k}   &= g\left(x_k, v_k\right)~\label{eq:gen_meas},
\end{align}
where $x_k$, $v_k$, and $y_k$ are vectors representing states, inputs, and output measurements at time $k$.
Additionally, $f$ is the dynamics model, and $g$ is the measurement model.
Linear state-space models are commonly used in control theory because they provide a tractable and simplified approach to analyzing and designing control systems. 
A linear state-space model utilizes matrices to approximate the $f$ and $g$ functions,
\begin{align}
    x_{k+1} & = Ax_k + Bv_k~, \label{eq:state}\\
    y_k & = Cx_k + Dv_k~, \label{eq:outp}
\end{align}
where $A$ and $B$ matrices are the state and input matrices, and $C$ and $D$ are the output and control pass-through matrices.
The linear state-space representation allows the application of powerful mathematical techniques such as eigenvalue analysis, frequency response analysis, and convex optimization algorithms (\eg Quadratic Programs for Model Predictive Control).
However, many physical systems exhibit nonlinear behavior, including in systems with time-varying parameters or when the system is operating in certain conditions (within steep saddle points).
Thus, using linear models can lead to inaccuracies or instability in controlling a nonlinear system.

A nonlinear dynamics model could address the deficiencies of a linear representation.
Specifically, a neural network could be used as the nonlinear function representing $f$ and $g$ in \cref{eq:gen_dynamics,eq:gen_meas}.
The previous section presented PINNs as an approach to solving PDEs, \cref{eq:pinn_gen}.
If the PDEs are a good approximation of the dynamics of a physical system, the PINN will approximate the system's state in the form \cref{eq:dyn_no_v}.
However, several issues arise.
First, the output of the PINN is a continuous representation of the system state $x=x(z,t)$.
A discrete representation could be defined by,
\begin{equation}
    x_k = \left[\mathcal{F}\left(z_0, t_k\right), \mathcal{F}\left(z_1, t_k\right),...,\mathcal{F}\left(z_M, t_k\right)  \right]^\intercal~,
\end{equation}
where $z_0,...,z_M$ represent the physical locations at which the system's state is defined, and $t_k$ is the discrete time.
Without loss of generality, we can assume that $t_k=k\Delta t$, where $k$ is an integer indexing time, and $\Delta t$ is a constant time-step size.
Therefore, the propagation of the state would be determined by,
\begin{align}
    x_0 &= \left[\mathcal{F}\left(z_0, 0\right), \mathcal{F}\left(z_1, 0\right),...,\mathcal{F}\left(z_M, 0\right)  \right]^\intercal~, \\
    x_1 &= \left[\mathcal{F}\left(z_0, t_1\right), \mathcal{F}\left(z_1, t_1\right),...,\mathcal{F}\left(z_M, t_k\right)  \right]^\intercal~, \\
    \vdots \nonumber \\
    x_N &= \left[\mathcal{F}\left(z_0, t_N\right), \mathcal{F}\left(z_1, t_N\right),...,\mathcal{F}\left(z_M,  t_N\right)  \right]^\intercal~.
\end{align}
While useful for systems without any inputs, additional modifications are needed to accommodate systems with varying control inputs, \ie \cref{eq:dyn_v},
\begin{equation}
    x_k = \left[\mathcal{F}\left(z_0, t_k, v_k\right), \mathcal{F}\left(z_1, t_k, v_k\right),...,\mathcal{F}\left(z_M, t_k, v_k\right)  \right]^\intercal~, \label{eq:dyn_no_x0}
\end{equation}
where $v_k$ is a vector representing control inputs at step $k$.
The control inputs can manipulate BCs, \eg the temperature or velocity at a pipe inlet, or modify jump conditions between locations, \eg pressure jump due to the presence of a pump.
A problem with \cref{eq:dyn_no_x0} is the assumption that the system's initial condition, $x(z,t=0)$, must always be the same.
For a dynamical system, an arbitrary final input value, $v_{k\rightarrow\infty}$, would result in a new settling point, $x_{k\rightarrow\infty}$.
If we want the state-space model to generalize, a form propagating forward from arbitrary $x_0$ is needed.
Therefore, the initial condition itself should be an input for the PINN,
\begin{equation}
    x_k = \left[\mathcal{F}\left(z_0, t_k, x_{0}, v_k\right), \mathcal{F}\left(z_1, t_k, x_{0}, v_k\right),...,\mathcal{F}\left(z_M, t_k, x_{0}, v_k\right)  \right]^\intercal~, \label{eq:dyn_w_x0}
\end{equation}
where $x_0$ is the initial condition.

\begin{figure}
    \centering
    \begin{tikzpicture}[scale=1.5, dot/.style={circle,fill,inner sep=1.5pt}]
    \node[] (init0) at (0, -4) {};
    \node[dot] (init1) at (1,-4) {};
    \node[above right] at (1,-4) {$x_{0,k}$};
    \node[dot] (init2) at (2,-4) {};
    \node[above right] at (2,-4) {$x_{0,k+1}$};
    \node[dot] (init3) at (3,-4) {};
    \node[above right] at (3,-4) {$x_{0,k+2}$};
    \node[dot] (init4) at (4,-4) {};
    \node[above right] at (4,-4) {$x_{0,k+3}$};
    \node[] (initN) at (5, -4) {};
    
    \node[] (state0) at (0, -2) {};
    \node[dot] (state1) at (1,-2) {};
    \node[above right] at (1,-2) {$x_k$};
    \node[dot] (state2) at (2,-2) {};
    \node[above right] at (2,-2) {$x_{k+1}$};
    \node[dot] (state3) at (3,-2) {};
    \node[above right] at (3,-2) {$x_{k+2}$};
    \node[dot] (state4) at (4,-2) {};
    \node[above right] at (4,-2) {$x_{k+3}$};
    \node[] (stateN) at (5, -2) {};
    
    \node[] (input0) at (0, -3) {};
    \node[dot] (input1) at (1,-3) {};
    \node[above right] at (1,-3) {$v_k$};
    \node[dot] (input2) at (2,-3) {};
    \node[above right] at (2,-3) {$v_{k+1}$};
    \node[dot] (input3) at (3,-3) {};
    \node[above right] at (3,-3) {$v_{k+2}$};
    \node[dot] (input4) at (4,-3) {};
    \node[above right] at (4,-3) {$v_{k+3}$};
    \node[] (inputN) at (5, -3) {};
    
    \draw[] (init0) -- (init1);
    \draw[] (state0) -- (state1);
    \draw[] (input0) -- (input1);
    \draw[] (init4) -- (initN);
    \draw[] (state4) -- (stateN);
    \draw[] (input4) -- (inputN);
    \foreach \x [count=\xi from 2] in {1,...,3} {
        \draw[-] (init\x) -- (init\xi);
        \draw[-] (state\x) -- (state\xi);
        \draw[-] (input\x) -- (input\xi);
    }

    \foreach \x in {1,...,4} {
        \draw[red] (init\x) -- (input\x);
    }
    
    \draw[dashed,->] (state0) -- (init1);
    \draw[dashed,->] (state4) -- (initN);
    \foreach \x [count=\xi from 2] in {1,...,3} {
        \draw[dashed,->] (state\x) -- (init\xi);
        \draw[blue,->] (input\x) -- (state\x);
    }
    \draw[blue,->] (input4) -- (state4);
    
\end{tikzpicture}
    \caption{A visualization of the nomenclature and their relationships.
             The red lines indicate variables needed at each step $k$, \ie the initial condition and input, $x_{0,k}$, and $v_k$.
             The blue arrows indicate the prediction of the next state, $x_k$, by the numerical solver or PSM.
             The dashed arrows indicate the aliasing of $x_k$ as $x_{0,k+1}$.
             } \label{fig:data_label}
\end{figure}
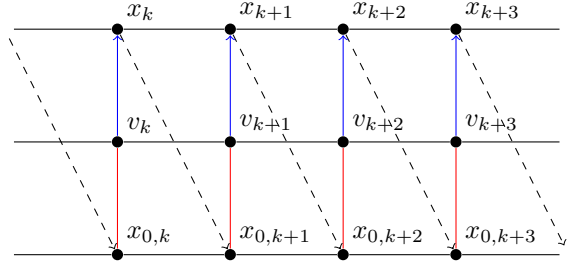

When a dynamical system is subject to an initial condition, the system's behavior in the short term is dominated by the initial state. 
However, as time passes, $k\gg1$, the impact of the initial condition on the system's current state becomes progressively smaller.
Therefore, a final modification to the formulation is needed,
\begin{equation}
    x_k = \left[\mathcal{F}\left(z_0, t_k, x_{0,k}, v_k\right), \mathcal{F}\left(z_1, t_k, x_{0,k}, v_k\right),...,\mathcal{F}\left(z_M, t_k, x_{0,k}, v_k\right)  \right]^\intercal~, \label{eq:dyn_w_x0k}
\end{equation}
where $x_{0_k}$ is the initial condition of the current time step.
Before \cref{eq:dyn_no_x0}, it was assumed that $t_k=k\Delta t$.
When this assumption is applied to \cref{eq:dyn_w_x0k}, over a collection of multiple experiments, the value of $t_k$ would arbitrarily correspond to combinations of the inputs, $(x_{0,k},v_k)$, and outputs, $x_k$.
From the perspective of training the neural net $\mathcal{F}$, there would be no relationship between $x_k$ and $t_k$.
However, $t_k$ is still an important variable to estimate the time derivatives needed for physics informing. 
The timescale at which $t_k$ does have a relationship between the inputs at outputs is the discrete time step, $\Delta t$.
The data would be separated into an initial condition at $t_k=0$, and the next step, $t_k=\Delta t$,
\begin{align}
    x_{0,k} &= \left[\mathcal{F}\left(z_0, 0, x_{0,k}, v_k\right), \mathcal{F}\left(z_1, 0, x_{0,k}, v_k\right),...,\mathcal{F}\left(z_M, 0, x_{0,k}, v_k\right)  \right]^\intercal,~\mathrm{and}, \label{eq:dyn_k} \\
    x_{k} &= \left[\mathcal{F}\left(z_0, \Delta t, x_{0,k}, v_k\right), \mathcal{F}\left(z_1, \Delta t, x_{0,k}, v_k\right),...,\mathcal{F}\left(z_M, \Delta t, x_{0,k}, v_k\right)  \right]^\intercal~. \label{eq:dyn_k+1}
\end{align}
The relationships between the inputs, $(x_{0,k},v_k)$, and outputs, $x_k$, in \cref{eq:dyn_w_x0k} is visualized in \cref{fig:data_label}.

The resulting neural network is,
\begin{equation}
    x\left(z_i, t_i, v_i, x_{0,i}\right) = \mathcal{F}\left(z_i, t_i, v_i, x_{0,i}\right)~, \label{eq:psm}
\end{equation}
predicting each component of \cref{eq:dyn_k,eq:dyn_k+1}.
To train this neural network, the datasets in \cref{eq:ds1,eq:ds2} are modified, 
\begin{align}
    \mathcal{X} &= \{(z_i, t_i, v_i, x_{0,i})_{i=1}^{N}\}~, \label{eq:ds3}\\
    \mathcal{Y} &= \{x(z_i, t_i, v_i, x_{0,i})_{i=1}^{N}\}~, \label{eq:ds4} 
\end{align}
where $v_i$ is the control input, and $x_{0,i}$ is the initial condition.
Formally, $v_i\in\mathbb{R}^p$ where $p$ is the number of control inputs, and $x_{0,i}\in\mathbb{R}^q$ where $q$ is the number of states modeled.
Using this formulation, the input space of the neural network increases by \O{10^{p+q}}.
However, this step is essential to allow network usage without relying on a heuristic approach, \eg a tabular database of multiple models where output is obtained by interpolating between $(v_i, x_{0,i})$.
Besides, this would circumvent a desire to show that the network can generalize well.
The trained network is referred to as a PSM.

\subsection{PSM architecture and training methodology} \label{subsec:archntrain}

To train a PSM, an ANN must be constructed.
The ANN will represent the operator, $\mathcal{F}$, in \cref{eq:meas_loss,eq:pinn_loss,eq:dyn_w_x0k}.
In past research, various PINN architectures have been explored, including Multi-Layer Perceptron (MLP), Convolutional Neural Networks (CNNs), and Recurrent Neural Networks (RNNs) \citep{cuomo_scientific_2022}.
For this study, we adopt an MLP architecture.
The intuition behind this choice is explained by contrasting an MLP's properties against other architectures.
Transport system dynamics are inherently Markovian, meaning that the state $x_{k+1}$ depends solely on the previous state, $x_k$, rather than a sequence of past states.
This characteristic makes a state-less architecture, such as an MLP, more suitable than an RNN for this application.
Additionally, CNNs are typically employed for tasks involving image or signal processing, where the input exhibits a grid-like structure and exhibits strong spatiotemporal correlations, such as multi-dimensional data.
However, in this study, the PSMs are designed to model one-dimensional transport, making an MLP architecture adequate for the task.

\begin{figure}
    \centering
    \includegraphics[width=\linewidth]{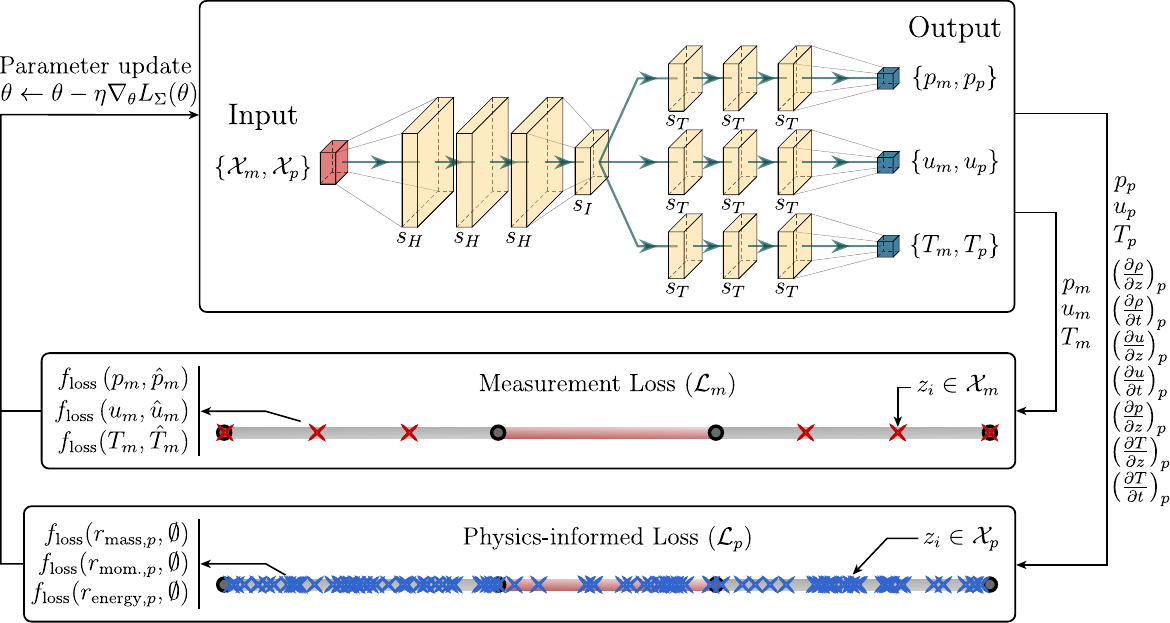}
    \caption{PSM architecture and training workflow.
    Top: The neural network architecture displaying relative sizes of the MLP layers.
    Bottom: Measurement and physics-informed losses use output from $\mathcal{F}\left(\mathcal{X}_m\right)$ and  $\mathcal{F}\left(\mathcal{X}_p\right)$, respectively.
    The combined losses are used to update the parameters of the neural network, $\theta$.}
    \label{fig:arch}
\end{figure}

Each \ith{i} layer of the MLP provides an output,
\begin{equation}
    y_j^i = \sigma^i\left(w_{jk}^i y_k^{i-1} + b_j^i\right)~,
\end{equation}
where $w_{jk}^{i}$ is the weight matrix, $b_j^i$ is a bias vector, and $\sigma^i$ is the activation function of the \ith{i} layer.
The vectors $y_j^i$ and $y_k^{i-1}$ are outputs of the \ith{i} and \ith{(i\!-\!1)} layers.
The sizes of the matrices and vectors (denoted by $j,~k$) will depend on the layer-wise size of $y$.
In this work, the ANN is built and trained with \pkg{PyTorch} \cite{paszke2019pytorch}. 
The architecture of the neural network and the training workflow is presented in \cref{fig:arch}.
The current configuration of the neural network is described next.
The activation function between all intermediate layers is the hyperbolic tangent.
It was chosen because it is nonlinear, is smooth, and has smooth derivatives.
After the input layer, there are three ``head'' layers of size $s_H$, followed by an ``intermediate'' layer of size $s_I$.
The output of the intermediate layer is sent to three separate ``tail'' layers of size $s_T$.
The layer sizes used are $(s_H,s_I,s_T)=(200,100,100)$.
There are three tails for each field variable: pressure, velocity, and temperature.
The intuition behind this architecture is that the head layers encode the initial condition and control input into a form used by the tail layers.
The tail layers then decode the information separately for each field variable.

As discussed previously, two data sets are used to train the PSM.
Measurement data from locations $z_i\in\mathcal{X}_m$, \cref{eq:ds1}, and physics-informing from collocation locations $z_i\in\mathcal{X}_p$, \cref{eq:ds_pinn}.
In the context of transport systems, the measurement locations are those where physical measurements of fluid flow exist, \eg thermocouples or pressure transducers.
On the other hand, collocation locations are arbitrarily chosen across the entire physical domain.

\subsubsection{Loss formulation}

A discussion is presented to explain the differences in the ``Measurement Loss'', \cref{eq:meas_loss}, and ``Physics-informed Loss'', \cref{eq:pinn_loss}, in \cref{fig:arch}.
In conventional PINN formulation, the objective is to solve for the spatiotemporal distribution of field(s) within the boundaries of a domain under the assumptions of a fixed initial condition and BC(s).
In general, it is anticipated that there are no measurements inside the spatial domain.
However, in transport systems, the direct measurements of pressure, temperature, and indirect measurement of velocity, are available through sensors (the type of sensor is dictated by the application, accuracy needed, and fluid properties) at multiple locations along the fluid transport path.
Thus, in the PSM formulation, the spatiotemporal distribution of field(s) \textit{between} the measurement locations must be solved.
In this setting, the measurement losses enforce initial conditions in \cref{eq:dyn_k} but also act as BCs in \cref{eq:dyn_k+1}.
On the other hand, the physics-informed losses are calculated at random locations.
The losses from these locations tune the coefficients of the PSM such that the PDE equations defined in \cref{eq:pinn_gen} are satisfied.
To tune the coefficients of the PSM, it is important to choose an appropriate loss function.

To train the network, the loss function, $f_\mathrm{loss}$ in \cref{eq:meas_loss,eq:pinn_loss}, must be defined.
The loss function quantifies the discrepancy between a target value and a model's prediction.
Regression problems in machine learning commonly use the Mean Squared Error (MSE) as the loss function. 
However, the Log-Cosh loss function, \cref{eq:lcl}, is utilized in this work. 
In both equations, $y_i$ is the predicted value, $\hat{y}_i$ is the target, and $N$ is the total number of samples in a batch.
\begin{equation}
    f_\mathrm{loss}\left(y_i,\hat{y}_i\right) = \frac{1}{N}\sum_{i=1}^N \log\left(\cosh\left(y_i-\hat{y}_i\right)\right)~\label{eq:lcl},
\end{equation}
The Log-Cosh loss function provides benefits over MSE by being less sensitive to outliers, having bounded gradients for numerical stability, ensuring more reliable convergence in optimization, and performing better with heavy-tailed noise distributions, allowing for quicker, more effective learning and accurate real-world data fitting \citep{wang2020comprehensive}.

Now that the definitions of the measurement loss, the physics-informed loss, and the loss function have been presented, a discussion on the weighting of the losses follows. 
The total loss forwarded to the optimizer per epoch is defined as,
\begin{equation}
    \mathcal{L}_\Sigma = \alpha\mathcal{L}_m + \beta\mathcal{L}_p ~ \big| ~ \alpha + \beta = 1~,\label{eq:loss_psm}
\end{equation}
where $\alpha$ and $\beta$ are coefficients that weigh the importance of the measurement and physics-informed losses, respectively.
The correct settings for $(\alpha,\beta)$ will depend on the degree of certainty in the measurements (assessable from the manufactured accuracy of sensors) and the confidence in the suitability of the PDEs' representation of the system's dynamics.
In this work, during training of the PSMs, we set $\alpha=0.5$ and $\beta=0.5$, equally weighting both losses.
However, further consideration is warranted when applying PSMs to a physical system.

\subsubsection{Scaling}

Scaling the input and output of a neural network is crucial when processing physical measurements of different units, scales, and ranges, such as pressure, temperature, and velocity.
This process facilitates the optimization of the neural network, enabling the model to converge more efficiently and quickly by ensuring that the gradients have a consistent magnitude across all input dimensions \citep{lecun2002efficient}. 
Unscaled inputs with disparate magnitudes can lead to elongated, narrow contours in the loss function landscape, resulting in slow and unstable convergence. 
In this work, min-max scaling is used, 
\begin{align}
z^* &= z/z_\mathrm{max}~,\\
t^* &= t/t_\mathrm{max}~, \\
p^* &= (p-p_\mathrm{min})/(p_\mathrm{max}-p_\mathrm{min})~, \\
u^* &= (u-u_\mathrm{min})/(u_\mathrm{max}-u_\mathrm{min})~, \\
T^* &= (T-T_\mathrm{min})/(T_\mathrm{max}-T_\mathrm{min})~, \\
\rho^* &= (\rho-\rho_\mathrm{min})/(\rho_\mathrm{max}-\rho_\mathrm{min})~,
\end{align}
for transforming position, time, pressure, velocity, temperature, and density, respectively.
The value for $z_\mathrm{max}$ equals the maximum path length for the liquid.
The value for $t_\mathrm{max}$ equals the discrete time-step between measurements, $\Delta t$ in \cref{eq:dyn_k+1}.
The minimum and maximum values for the pressure, velocity, temperature, and density are based on the ranges of data that are collected from experiments.

\subsubsection{Training procedure}

The PSM training procedure is listed in \cref{alg:PSM}.
During each epoch, the weights are updated from a fixed measurement dataset.
Whereas the collocation points are updated at each epoch.
A step learning rate scheduler was used, with a rate decay of 0.5 every 50 epochs.
All models were trained for 500 epochs with a batch size of 2048. 
These settings attempt to stabilize the training of the neural network and to decrease the components of $\mathcal{L}_{\Sigma}$ by 3-5 orders of magnitude. 

\begin{algorithm}
\caption{Training a PSM}
\label{alg:PSM}
\begin{algorithmic}[1]
\State \textbf{Input:} Measurement data $(\mathcal{X}_m, \mathcal{Y})$, Governing equations $G$, Loss coefficients $(\alpha, \beta)$, Number of epochs $N$, Batch size $B$, Learning rate scheduler $\mathcal{S}(\cdot)$
\State \textbf{Initialize:} Network $\mathcal{F}$ with random weights $\theta$
\For{$n=1,2,\dots,N$}
    \State Update learning rate $\eta \leftarrow \mathcal{S}(n)$
    \For{each mini-batch $(\mathcal{X}_{m,i}, \mathcal{Y}_i) \subset (\mathcal{X}_m, \mathcal{Y})$ with size $B$}
        \State Compute predictions $\hat{\mathcal{Y}}_i = \mathcal{F}(\mathcal{X}_{m,i})$
        \State Compute measurement loss $\mathcal{L}_{m}(\hat{\mathcal{Y}}_i, \mathcal{Y}_i)$ using \cref{eq:meas_loss}
        \State Randomly sample $B$ collocation points $\mathcal{X}_{p,i}$, and extract $(x_{0,i}, v_i)$ from $\mathcal{X}_{m,i}$ 
        \State Compute the residual values $\mathcal{R}_i = G(\mathcal{F}, \mathcal{X}_{p,i}, x_{0,i}, v_i)$
        \State Compute physics-informed loss $\mathcal{L}_{p}(\mathcal{R}_i,\emptyset)$ using \cref{eq:pinn_loss}
        \State Compute total loss $\mathcal{L}_{\Sigma} = \alpha \mathcal{L}_{m} + \beta \mathcal{L}_{p}$
        \State Update network weights using optimizer: $\theta \leftarrow \theta - \eta \nabla_\theta L_{\Sigma}(\theta)$
    \EndFor
\EndFor
\State \textbf{return} Trained PSM model $\mathcal{F}$
\end{algorithmic}
\end{algorithm}

\section{Results: applying PSMs for transport in a Heated Channel} \label{sec:res_ch}

\begin{figure}
    \centering
    \begin{tabular}{@{}cc@{}}
        \begin{adjustbox}{valign=c}
            \includegraphics[scale=0.85]{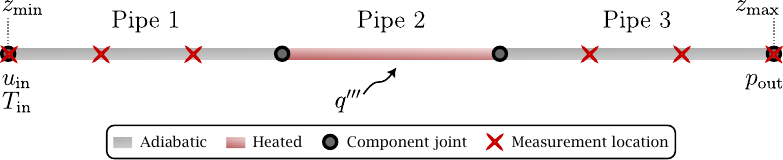}
        \end{adjustbox} &
        \begin{tabular}[c]{@{}|cl@{}}
             $p_\mathrm{out}$ & \SI{135.6}{\kilo\pascal}   \\
             $u_\mathrm{in,min}$ & \SI{0.549}{\meter\per\second}   \\
             $u_\mathrm{in,max}$ & \SI{0.749}{\meter\per\second}   \\
             $T_\mathrm{in,min}$ & \SI{531.5}{\celsius}   \\
             $T_\mathrm{in,max}$ & \SI{611.5}{\celsius} \\
        \end{tabular}
    \end{tabular}
    \caption{Configuration of the pipes in series.
    Left: Layout of the pipes and identifications of the locations of imposed BCs and heat source term.
    Right: A list of the fixed outlet pressure BC, and ranges for the inlet BCs.
    }
    \label{fig:exp_pipe}
\end{figure}

\begin{figure}
    \centering
    \begin{tabular}{cc}
       Control Input Permutations  & Exp. 8 Numerical Solution  \\
       \includegraphics[scale=0.7]{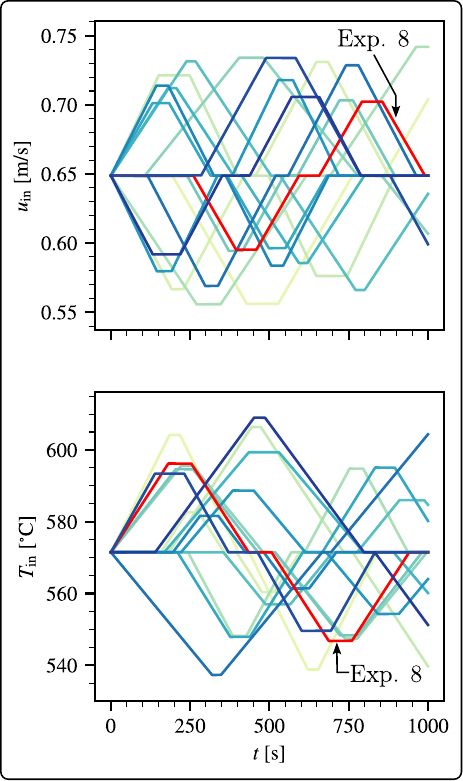} &
       \includegraphics[scale=0.7]{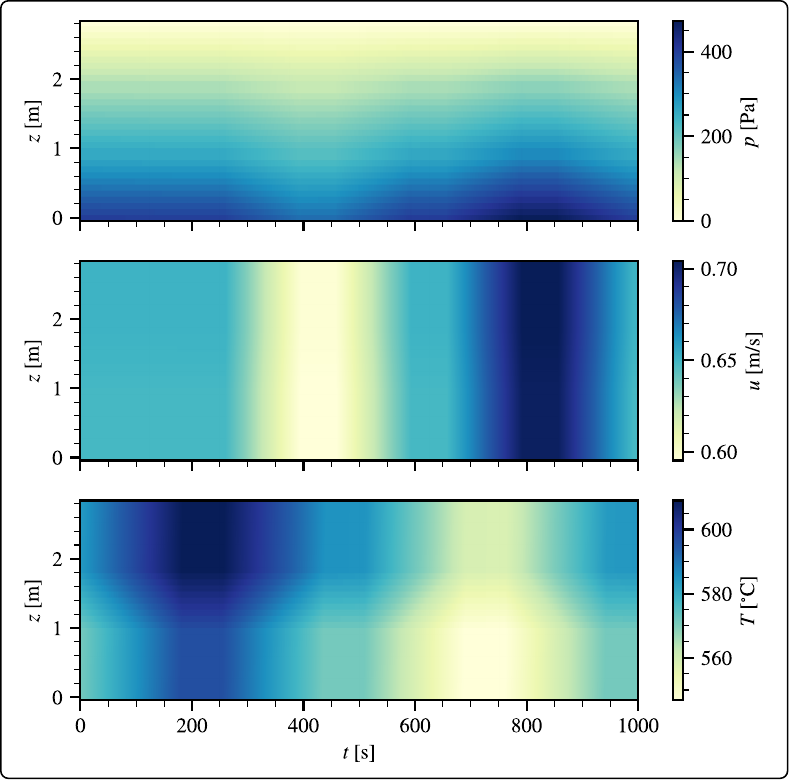}
    \end{tabular}
    \caption{An overview of the training dataset for the heated channel configuration.
    Left: Velocity and temperature BCs at the inlet that are manipulated in time.
    The experiment chosen to display the solution is annotated ("Exp. 8").
    Right: Numerical solution of all field variables over the entire spatiotemporal domain.}
    \label{fig:exp_results}
\end{figure}

Two experiments are designed to investigate PSMs' performance for transport phenomena.
The first experiment is a system of three conduits in series (referred to as pipes).
The first and last pipes will be adiabatic and \SI{1.0}{\meter} long.
The middle pipe will be heated with a constant volumetric heat source of $q'''=\SI{50.0}{\mega\watt\per\meter\cubed}$, and \SI{0.8}{\meter} long.
All pipes have a total cross sectional area of $\SI{0.449}{\meter\squared}$ and a hydraulic diameter of \SI{2.972}{\milli\meter}.
No solids are modeled -- only the internal flow area.
Each pipe is discretized into ten elements.
The inlet of the first pipe will have varying velocity and temperature BCs,
\begin{align}
    u(z=z_\mathrm{min}, t) &= u_\mathrm{in}(t)~, \\
    T(z=z_\mathrm{min}, t) &= T_\mathrm{in}(t)~.
\end{align}
The trajectory of the BCs is defined by linear ramps that vary in ramp rate, maximum/minimum amplitude, and rest time between manipulations. 
These parameters are varied to encourage regularization of the PSM.
Additionally, the outlet of the third pipe will be a constant pressure BC,
\begin{equation}
    p(z=z_\mathrm{max}, t) = p_\mathrm{out}. \
\end{equation}
In this work, the pressure BCs are presented as absolute pressure; however, the pressure field is presented as gage pressure.
The experiment configuration is visualized in \cref{fig:exp_pipe}.
A visualization of the evolution of the field variables is presented in \cref{fig:exp_results}.
The PSM will be tasked with predicting these spatiotemporal distributions by learning from a few measurement locations and the approximate PDEs of the system. 
Hereafter, this experiment will be referred to as the ``heated channel'' configuration.

To close the transport equations, \cref{eq:mass,eq:mom,eq:nrg}, the thermophysical properties of the fluid must be defined.
This work utilizes a molten salt, \ce{LiF-BeF_2}, known colloquially as `flibe.' 
The reason is that flibe is the leading candidate for the primary working fluid of next-generation molten salt reactor designs currently under development \citep{rehm2023advanced}.
The density and specific heat capacity of flibe are defined by \citep{romatoski2017fluoride},
\begin{align}
    \rho = 2413 - 0.488 T~[\SI{}{\kilo\gram\per\meter\cubed}]~, \label{eq:rho} \\
    C_p  = 2414 ~[\SI{}{\joule\per\kilo\gram\per\kelvin}]~,
\end{align}
where $T$ is the fluid temperature in Kelvin.
These correlations are adopted to calculate the physics-informed loss in \cref{eq:pinn_loss}.


\subsection{Evaluation with test data}\label{subsec:results-ch}

The results for the heated channel are discussed in this section.
A sample of the control input permutations and numerical solution of the field variables were presented in \cref{fig:exp_results}.
In this setting, the PSM must accurately propagate the system's state, $x_k$, from a given initial condition, $x_{0,k=0}$, and input trajectory, $v_{k=0...N}$.
This problem is a classical advection problem where the inlet boundary conditions are transported through the pipe incur: pressure losses due to friction from the pipe, modifications in mass flow rate (velocity) due to boundary conditions, and changes in temperature due to energy deposition.

\begin{figure}
    \centering
    \begin{tabular}{p{0.18\linewidth}p{0.4\linewidth}p{0.35\linewidth}}
         \centering Control Input & \centering Field values at fixed positions & \centering Field values at fixed times
    \end{tabular}
    \includegraphics[width=\linewidth]{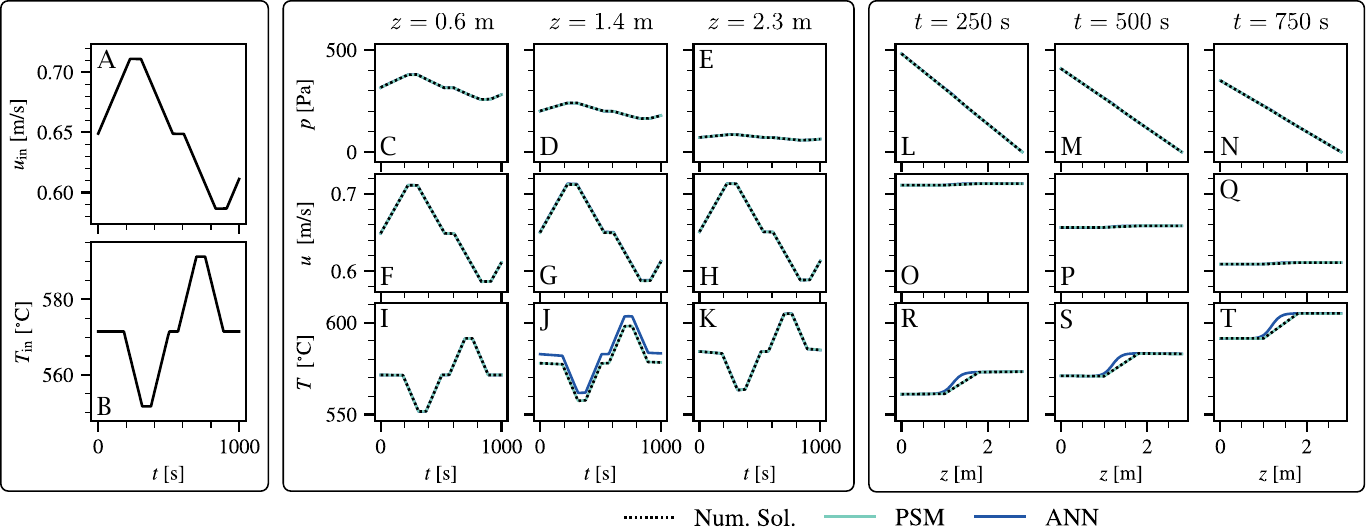}
    \caption{Contrasting the performance of PSM and ANN models in predicting the evolution of pressure, velocity, and temperature for the heated channel.
        Subplots A-B: control input setting from the test dataset.
        Subplots C-K: field temporal evolution at fixed positions.
        Subplots L-T: field spatial distribution at fixed times.
    }
    \label{fig:res_ch}
\end{figure}

In \cref{fig:res_ch}, a rollout of a single input, selected from the \textit{test} dataset, \ie data that is not used to train the models, is presented.
Subplots A and B show the input trajectory of the velocity and temperature at the inlet, $u_\mathrm{in}$ and $T_\mathrm{in}$.
Subplots C-K show pressure, velocity, and temperature at fixed positions \vs time.
Subplots L-T show the fields at fixed times \vs position.
In subplots C-T, results from the numerical solver (``Num. Sol.''), the PSM model, and an Artificial Neural Network (ANN) model are displayed.
The PSM model is trained with \cref{eq:loss_psm}, which contains both measurement and physics-informed loss, \cref{eq:meas_loss,eq:pinn_loss}, respectively.
Whereas the ANN model uses the same neural network architecture as the PSM but is trained only with the measurement loss, \cref{eq:meas_loss}.
Thus, the ANN results represent the classical, purely data-driven approach to modeling. 
In contrast, the PSM results combine knowledge from data with a priori knowledge of the system's dynamics in the form of transport PDEs.

\begin{figure}
    \centering
    \begin{tabular}{p{0.030\linewidth}p{0.1687\linewidth}p{0.0745\linewidth}p{0.17\linewidth}p{0.055\linewidth}}
         & \centering PSM & & \centering ANN &  \\
    \end{tabular}
    \includegraphics[scale=0.8]{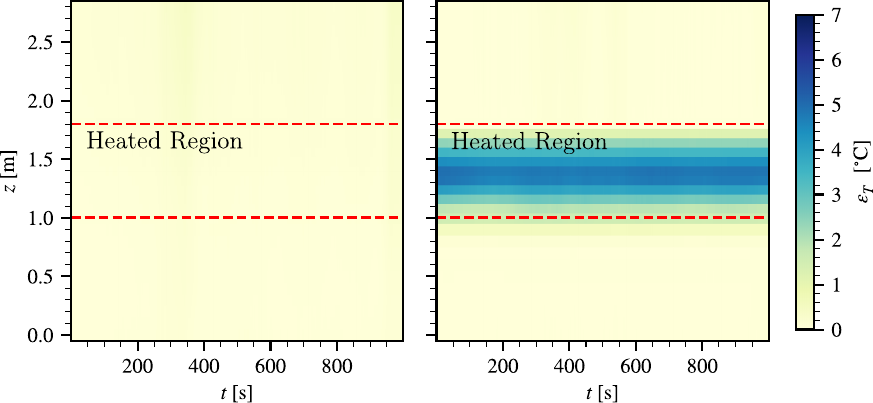} 
    \caption{Comparison of root-mean-square error (RMSE) in the prediction of temperature for PSM (left) and ANN (right) models.
             The RMSE values are averaged over all test datasets.
             The heated region is present between the dashed red lines.}
    \label{fig:temp_comp}
\end{figure}

The ANN appears to perform on par with the PSM.
Comparing the pressure and velocity spatiotemporal predictions, indeed, the ANN does perform as well as the PSM.
However, comparing the temperature distributions, there is a significant error introduced by the ANN.
The errors occur at $1.0 \leq z \leq \SI{1.8}{\meter}$, \ie the heated pipe's location.
This observation is more explicitly visualized in \cref{fig:temp_comp}, which compares spatiotemporal error in predicting temperature averaged over all test datasets.
The PSM significantly outperforms the ANN in the prediction of the temperature field.

Why does the ANN fail to accurately predict the temperature field only?
This question can be answered by understanding the underlying physics and the sensor measurement locations.

The advection of pressure and velocity depends on the frictional losses and the inlet mass flow rate.
Because the friction factor and pipe geometry are constant, we expect a constant pressure gradient with respect to position.
This is displayed in Subplots L-N in \cref{fig:res_ch}.
The measurement of the pressure and velocity fields occurs in the first and last unheated pipes.
Therefore, the ANN has to simply interpolate between the measurement locations of pressure before and after the heated section to correctly estimate the pressure field and be able to modify the gradient according to the change in inlet velocity mass flow rate.
The latter is provided as an input ($v_k$) to the ANN.
The results show that the ANN model can accomplish these tasks well for the pressure field.

For the velocity field, we expect the spatial distribution to be nearly constant, except after the heated region.
There is a change in the density because of the heating (\cref{eq:rho}), which causes the velocity to change in order to conserve mass (\cref{eq:mass}).
This causes a small error in ANN prediction, noticeable upon close inspection of Subplots O-Q in \cref{fig:res_ch}.
Therefore, the performance for velocity is still acceptable.

However, for the temperature field, there is a large discrepancy between ANN and PSM performance within the heated region.
The temperature field is affected by energy deposition ($q'''$ in \cref{eq:nrg}), and inlet BCs.
When $q'''=0$, there is no gradient in temperature.
When $q'''=c$, a constant, there is a linear gradient in temperature.
The PSM correctly predicted both situations, as shown in Subplots I-K and R-T in \cref{fig:res_ch}.
However, the ANN has no knowledge of $q'''$, and thus it interpolates between the sensor readings. 
Therefore, the ANN model can only correctly predict temperature distributions in regions where $q'''=0$, \ie there is no temperature gradient and any modification is provided directly by the inlet BC.

\subsection{Training with noisy measurements}\label{subsec:noise-ch}

\begin{figure}
    \centering
    \includegraphics[width=\linewidth]{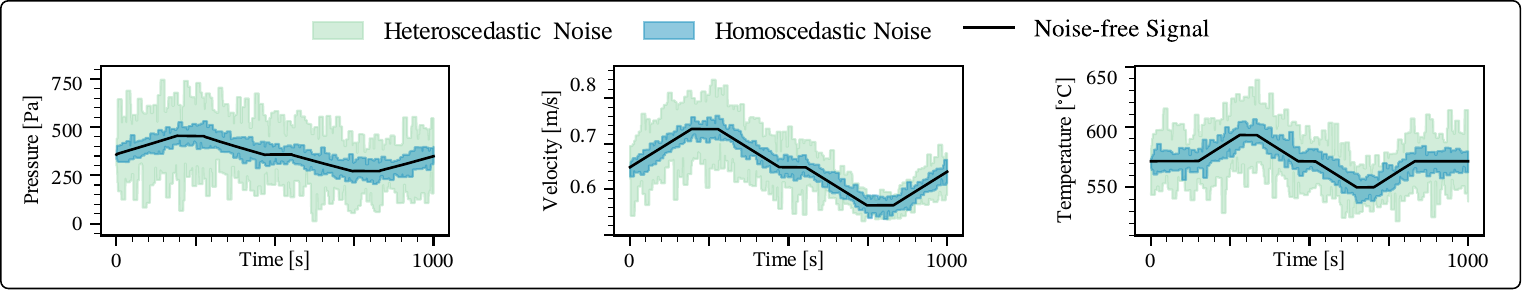}
    \caption{Example of the manipulation of sensor measurements for all fields.
             During training, the signals are manipulated randomly for each batch.}
    \label{fig:noise}
\end{figure}

The data from sensor measurements in a physical plant are noisy.
The noise stems from inherent randomness during fluid transport \& signal measurement itself.
It is important to quantify the performance of PSMs in such environments.
The introduction of homoscedastic and heteroscedastic noise is studied.
The former is introduced by,
\begin{equation}
x_\mathrm{noisy} = x + \sigma \cdot \epsilon~,\label{eq:homo-noise}
\end{equation}
where $x_\mathrm{noisy}$ is the noisy signal, $x$ is the noise-free signal, $\sigma$ is the noise standard deviation, and $\epsilon$ is a random noise vector sampled from a standard normal distribution.
This type of noise model scenarios where the variance is constant across all magnitudes.
Heteroscedastic noise is introduced by,
\begin{equation}
x_\mathrm{noisy} = x + \sqrt{|x| \cdot \xi} \cdot \epsilon~,\label{eq:hetero-noise}
\end{equation}
where $\xi$ is the variance factor.
The variance factor is a parameter that controls the relationship between the signal's magnitude and the added noise's variance.
Thus, this type of noise model scenarios where the noise variance is not constant but depends on the underlying signal's magnitude.
The noisy signals are visualized in \cref{fig:noise}.
During training, the signals are manipulated randomly for each batch.
Noise is added to the sensor measurements and the initial conditions ($x_k$ and $x_{0,k}$, respectively, in \cref{eq:dyn_k+1}).
It is assumed that the system's actuation is not noisy (\ie the control signals, $v_k$ are noise-free).

The results from introducing noise are discussed next.
The training procedure remains the same, except that artificial noise is added during each batch.
In \cref{tab:noise}, the results are tabulated for both types of noise, and varying magnitudes of noise factors ($\sigma$ in \cref{eq:homo-noise}, $\xi$ in \cref{eq:hetero-noise}).
First, inspecting the mean RMSE values, the PSM has a significant advantage in predicting velocity and temperature fields.
There is a lower advantage for the pressure fields.
This might be because the nominal RMSE values for pressure are already very low when contrasted with the pressure range ($\Delta P$) across all experiments.
Inspecting the maximum RMSE values, the PSM retains a significant advantage in predicting velocity and temperature fields, with some anomalies noted for $\sigma=0.05$ and $\xi=0.05$.
The PSM's advantage for the pressure field is relatively moderate, likely because it has a simple spatiotemporal relationship in the current experimental setting.
To summarize, the results show that even when noise is added to the signals, PSMs have a significant advantage over a purely data-driven model.

\begin{table}[]
    \centering
    \begin{tabular}{l c c c c c c}
    & $\varepsilon_{P,\mathrm{ANN}} [\SI{}{\pascal}]$ & $\varepsilon_{u,\mathrm{ANN}} [\SI{}{\milli\meter\per\second}]$ & $\varepsilon_{T,\mathrm{ANN}} [\SI{}{\celsius}]$ & $\frac{\varepsilon_{P,\mathrm{PSM}}}{\varepsilon_{P,\mathrm{ANN}}} [\%]$ & $\frac{\varepsilon_{u,\mathrm{PSM}}}{\varepsilon_{u,\mathrm{ANN}}} [\%]$  & $\frac{\varepsilon_{T,\mathrm{PSM}}}{\varepsilon_{T,\mathrm{ANN}}} [\%]$ \\
    \hline
    \multicolumn{7}{|c|}{Mean} \T\B\\
    \hline\T
    Noise-free      & 0.51	& 0.17	& 1.01 &	\cf{58}     & \cf{52}	& \cf{6 } \\
    Homos. $\sigma=0.01$ & 1.25	& 0.25	& 1.15 &	\cf{12}	    & \cf{25}	& \cf{3 } \\
    Homos. $\sigma=0.05$ & 0.46	& 0.22	& 0.98 &	\cf{143}    & \cf{35}	& \cf{17} \\
    Homos. $\sigma=0.10$ & 0.83	& 0.23	& 1.01 &	\cf{91}	    & \cf{51}	& \cf{28} \\
    Heteros. $\xi=0.01$ & 0.71	& 0.25	& 1.15 &    \cf{94}	    & \cf{41}	& \cf{24} \\
    Heteros. $\xi=0.05$ & 0.94	& 0.27	& 1.10 &    \cf{94}	    & \cf{68}	& \cf{27} \\
    Heteros. $\xi=0.10$ & 1.07	& 0.34	& 0.97 &    \cf{89}	    & \cf{63}	& \cf{33} \\
    \hline
    \multicolumn{7}{|c|}{Maximum} \T\B\\
    \hline\T
    Noise-free  & 4.82	&   1.25	&   5.81	&   \cf{113}  &  \cf{52}  & \cf{31} \\
    Homos. $\sigma=0.01$ & 8.97	&   1.24	&   6.08	&   \cf{10	} &  \cf{43 } & \cf{13 }\\
    Homos. $\sigma=0.05$ & 6.17	&   1.24	&   7.00	&   \cf{53	} &  \cf{104} & \cf{14 }\\
    Homos. $\sigma=0.10$ & 6.49	&   2.38	&   6.87	&   \cf{78	} &  \cf{35 } & \cf{27 }\\
    Heteros. $\xi=0.01$ & 4.80	&   1.34	&   7.70	&   \cf{81	} &  \cf{51 } & \cf{19 }\\
    Heteros. $\xi=0.05$ & 5.47	&   1.42	&   6.54	&   \cf{95	} &  \cf{120} &	\cf{35 }\\
    Heteros. $\xi=0.10$ & 6.31	&   1.97	&   5.46	&   \cf{76	} &  \cf{69 } & \cf{38 }\\
    \hline
    \multicolumn{7}{|c|}{Range across all experiments: $\Delta P = \SI{531}{\pascal}$, $\Delta u = \SI{200.0}{\milli\meter\per\second}$, $\Delta T = \SI{92.6}{\celsius}$}  \T\B\\
    \hline\T
    \end{tabular}
    \caption{Comparison of the mean and maximum RMSE for field variables using the ANN across all test datasets.
             The relative RMSE error for PSMs is displayed in the last three columns: the values are color-coded to highlight significant performance improvement when below 75 \% and a degradation in performance if above 100 \%.
             The range of field values over all experiments is displayed at the bottom.}
    \label{tab:noise}
\end{table}

\subsection{PSM Capability: Model-based Control}\label{subsec:results-control}

In the previous section, the PSM architecture was shown to accurately approximate the spatiotemporal distribution of all the fields.
Because the PSM is an end-to-end differentiable model, we can calculate the gradients for inputs and outputs using auto differentiation.
In this section, the capability of a PSM to enable supervisory control of a nonlinear system is demonstrated.
The Reference Governor (RG) supervisory control scheme is adopted to demonstrate this capability.
The RG is a model-based scheme that enforces pointwise-in-time constraints on a system by modifying the reference input setpoints \citep{Garone2017}.

The RG algorithm is summarized next.
At each time-step $k\in\mathbb{Z}_+$, the RG receives a reference input, $r_k\in\mathbb{R}^m$, where $m$ is the number of inputs.
Depending on the system's current state, expected evolution, and imposed constraints, an admissible input $v_k\in\mathbb{R}^m$,
\begin{equation}
    v_k = v_{k-1} + \kappa_k \left(r_k - v_{k-1}\right)~,
    \label{eq:rg1}
\end{equation}
is obtained.
The admissible input is then sent to lower-level controllers of the system (\eg a PID controller).
In the Scalar RG (SRG) formulation, $\kappa_k\in[0,1]$ is a \textit{scalar} that governs admissible changes to the inputs, such that a complete rejection, $v_k=v_{k-1}$, acceptance, $v_k=r_k$, or an intermediate change, $v_{k-1}\leq v_k\leq r_k$, is possible.

Next, how the SRG algorithm determines $\kappa_k$ is summarized.
The RG uses a linear state-space representation, \cref{eq:state,eq:outp}, and then imposes constraints on the output variables, $y_k\in Y$, where a set of linear inequalities defines $Y$.
Then, a construct referred to as the Maximal Output Admissible Set ($O_\infty$) is defined -- it is the set of all $x_k$, and constant input $\Tilde{v}$, such that,
\begin{equation}
    O_\infty = {(x_k, \Tilde{v}) : y_{t+k}\in Y, v_{t+k}=\Tilde{v}, \forall k \in \mathbb{Z}^+}~.
    \label{eq:moas}
\end{equation}
The $k\rightarrow \infty$ assumption of $O_\infty$ is generally relaxed to a suitably large finite horizon, $T$.
Thus, at each time-step $k$, the SRG algorithm determines the admissible $\Tilde{v}$, and therefore $\kappa_k$, such that the system remains in $O_\infty$.
The $O_\infty$ is constructed by a roll-out of \cref{eq:state,eq:outp}, detailed further in \citep[\S 2]{Garone2017}.
For a fixed set of state-space matrices and constraints, the matrices needed to construct $O_\infty$ can be precomputed and stored for each RG evaluation.

A drawback of the SRG formulation is that $\kappa_k$ is a scalar.
If $v_k-r_k$ is multi-dimensional and $\kappa_k<1$, input movement in all dimensions is bounded.
In this work, we utilize the Command Governor (CG), a variant of the RG which selects $v_k$ by solving the quadratic program,
\begin{align}
    v_k = &\argmin_{v_k} ||v_k - r_k ||^2_Q~, \nonumber \\
    \mathrm{s.t.} \hspace{6pt} &(x_k, v_k=\Tilde{v})\in O_\infty~, \label{eq:cg}
\end{align}
where $Q$ is a positive definite matrix signifying the relative importance of each $v_k$ component.
The quadratic program is solved by the \texttt{CVXPY} library \cite{diamond2016cvxpy}.
Using the CG, constraints can be enforced by manipulating $v_k$ arbitrarily in $\mathbb{R}^m$, constrained by $O_\infty$.

To utilize the CG, a linear representation of the system is needed.
This work assumes full-state feedback, $C=\mathbb{I}$, and no control pass-through, $D=\mathbf{0}$.
Because the PSM is end-to-end differentiable, we can approximate the nonlinear model, \cref{eq:gen_dynamics}, as a linear representation using Jacobian linearization,
\begin{align}
    x_{k+1} &\approx x_{00} + A\delta x_k + B \delta v_k~,\\
    \delta x_k &= x_k - x_{00}~\label{eq:lpx},\\
    \delta v_k &= v_k - v_{00}~\label{eq:lpv},
\end{align}
where $x_{00}$ and $v_{00}$ are state and input values at the linearization point.
The matrices $A$, and $B$ are approximated by the Jacobians of $\mathcal{F}$, \cref{eq:psm}, at the linearization point,
\begin{align}
    A &= \left[ \frac{\partial\mathcal{F}_{x_{k+1}}}{\partial x_{k,0}},\frac{\partial\mathcal{F}_{x_{k+1}}}{\partial x_{k,1}},...,\frac{\partial\mathcal{F}_{x_{k+1}}}{\partial x_{k,N}} \right]_{(x_{00},v_{00})}^\intercal~, \label{eq:jacA} \\
    B &= \left[ \frac{\partial\mathcal{F}_{x_{k+1}}}{\partial v_{k,0}},\frac{\partial\mathcal{F}_{x_{k+1}}}{\partial v_{k,1}},...,\frac{\partial\mathcal{F}_{x_{k+1}}}{\partial v_{k,M}} \right]_{(x_{00},v_{00})}^\intercal~ \label{eq:jaqB},
\end{align}
where $N$ and $M$ are the number of states and inputs.
In summary, the RG was introduced as the framework to enforce constraints, followed by a specific version, the CG, which can manipulate multiple inputs to enforce constraints on a system.
Then, the approach to utilize the PSM to approximate the linear state-space matrices required by the CG was introduced.
The algorithm to utilize this procedure for supervisory control is listed in \cref{alg:NCG}.

\begin{algorithm}
\caption{Sequential Neural Command Governor (NCG) using PSMs}
\label{alg:NCG}
\begin{algorithmic}[1]
\State \textbf{Input:} PSM model $\mathcal{F}$, Requested set point trajectory $r$, Time-dependent constraints on states $Y_k$, Input matrix $Q$, Matrices $A$ and $B$ update interval $\gamma$
\State \textbf{Initialize:} Initial condition of system $x_0$, Iteration counter $i \gets 0$
\For{$k = 1$ to $\text{length}(r)$}
    \If{$k\mod\gamma = 0$}
        \State Obtain matrices $A$ and $B$ by Jacobian Linearization of $\mathcal{F}$ at state $x_k, v_k$
        \State Update the $O_\infty$ matrices with the current linear state-space model
    \EndIf
    \If{Constraint update}
        \State Update the $O_\infty$ matrices with the current $Y_k$
    \EndIf
    \State Formulate the quadratic program in \cref{eq:cg} with inputs $r_k, x_k$
    \State Solve the quadratic program to obtain the admissible set point $v_k$
    \State \textbf{Return} Admissible set point $v_k$ to controlled system
    \State Update state of system $x_k \gets x_{k+1}$
\EndFor
\end{algorithmic}
\end{algorithm}

\begin{figure}
    \centering
    \includegraphics[scale=0.7]{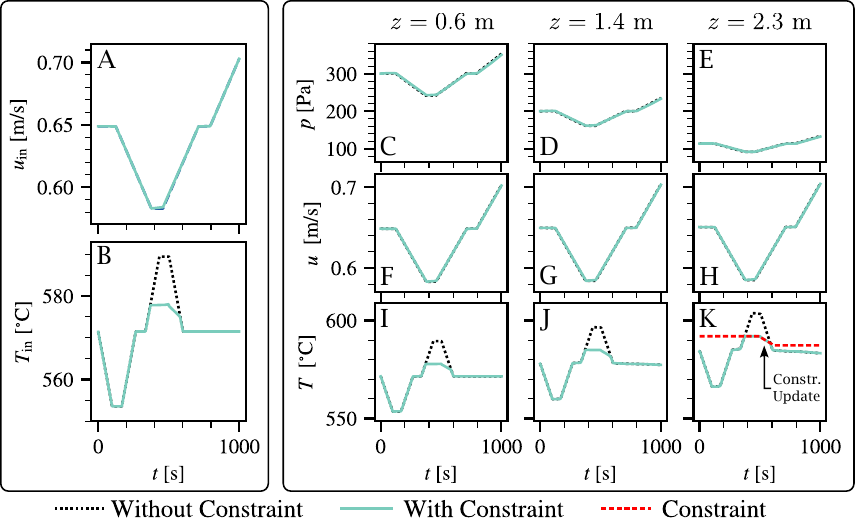}
    \caption{Demonstration of the NCG algorithm.
    The PSM instantiates the models needed to utilize the supervisory control algorithm.
    The numerical solver is used as the environment with which the NCG algorithm interacts.
    Subplots A-B: Control input permutations. 
    The input value without constraints is the reference input trajectories for the NCG algorithm.
    Subplots C-K: Field values at fixed locations.
    Subplot K: A constraint is assigned at this location and manipulated as a function of time.}
    \label{fig:sqp}
\end{figure}

Next, results demonstrating \cref{alg:NCG} are presented.
In \cref{fig:sqp}, an input permutation from the test dataset is chosen randomly to manipulate the inlet BCs.
The variation of the temperature BC, $T_\mathrm{in}$, causes a significant temperature rise after the heated channel, observable by comparing subplots I-K.
It is desirable to constrain excessive increases in temperature to avoid damage to equipment or mitigate corrosion. 
Thus, a temperature constraint is assigned at a location after the heated channel, $z=\SI{2.3}{\meter}$, shown in subplot K.
The NCG algorithm admits input changes to enforce these constraints, shown in subplots A and B.
The results demonstrate that the approach is successful in meeting the enforced constraints.
There are a few key takeaways.
First, the NCG algorithm independently manipulates each input value, minimizing the distance from the requested input trajectories, \ie \cref{eq:cg}.
Second, a time-dependent constraint is demonstrated (annotated in subplot K).
Accommodating time-dependent constraints enables flexibility during operation -- it is envisioned that, as part of a comprehensive autonomous operation framework, an external algorithm could update the constraints enforced on a system depending on its health condition.

It is important to note that in this demonstration, the PSM is only used to update the $A$ and $B$ matrices -- the numerical solver is used as the environment with which the NCG algorithm interacts.
The accuracy of Jacobean linearization is generally high near the linearization point, $(x_{00},v_{00})$, but can decrease as the system moves away from this point.
It is most effective for systems that exhibit mild nonlinearity around the operating point.
Therefore, when implemented in practice, the matrix update interval, $\gamma$ in \cref{alg:NCG}, must be chosen carefully.
Lastly, a model-based control scheme may have additional stability assumptions.
For the RG, it is assumed that the system being controlled is asymptotically stable, \ie the matrix $A$ is Schur \cite{Garone2017}.
This can be verified through eigenvalue analysis of the linearized matrix.

\section{Results: applying PSMs for transport in a Cooling System}\label{sec:res_loop}

\begin{figure}
    \centering
    \begin{tabular}{@{}cc@{}}
        \begin{adjustbox}{valign=c}
            \includegraphics[scale=0.85]{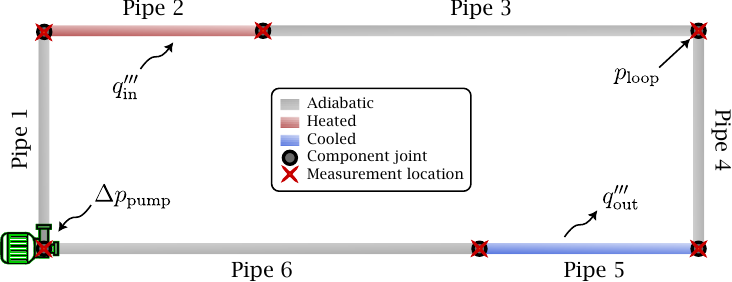}
        \end{adjustbox} &
        \begin{tabular}[c]{@{}|cl@{}}
             $q'''_\mathrm{in/out,min}$ & \SI{45.0}{\mega\watt\per\meter\cubed}   \\
             $q'''_\mathrm{in/out,max}$ & \SI{55.0}{\mega\watt\per\meter\cubed}   \\
             $p_\mathrm{loop}$ & \SI{100.0}{\kilo\pascal} \\
             $\Delta p_\mathrm{min}$ & \SI{1125}{\pascal}   \\
             $\Delta p_\mathrm{max}$ & \SI{1875}{\pascal}   \\
        \end{tabular}
    \end{tabular}
    \caption{Configuration of pipes in a loop.
    Left: The imposed pump BC location and heat source/sink terms.
    Right: A list of the ranges for the pump pressure jump, heat source/sink terms, and loop pressure setting.
    }
    \label{fig:exp_loop}
\end{figure}

In addition to the heated channel configuration, a ``cooling system'' configuration was adopted.
A series of six pipes are arranged in a loop.
All pipes have a total cross sectional area of $\SI{0.449}{\meter\squared}$ and a hydraulic diameter of \SI{2.972}{\milli\meter}.
No solids are modeled -- only the internal flow area.
All pipes are \SI{1.0}{\meter} in length, except for the pipes immediately after the heated and cooled sections, which are \SI{2.0}{\meter} long.
The pipes are discretized into 10 elements per meter.
A pump provides the head necessary to drive the fluid flow.
In addition to a volumetric heat source, a volumetric heat sink is also present.
This configuration is commonly found in advanced cooling systems, such as those used in high-performance electronic devices, nuclear reactors, and aerospace applications.
The loop configuration is visualized in \cref{fig:exp_loop}.
There are two independent control inputs,
\begin{align}
    q'''_\mathrm{in} &= q'''_\mathrm{in}(t)~, \\
    q'''_\mathrm{out} &= -q'''_\mathrm{in}(t)~, \label{eq:cooling}\\
    \Delta p_\mathrm{pump} &= \Delta p_\mathrm{pump}(t)~,
\end{align}
where $q'''_\mathrm{in}$ is the volumetric heat generation rate, $q'''_\mathrm{out}$ is the volumetric cooling rate, and $\Delta p_\mathrm{pump}$ is the pressure head provided by the pump.
A control strategy is adopted that synchronizes the heating and cooling rates to increase/decrease the system's heat generation/cooling capability while maintaining constant temperatures throughout the loop.
Both independent control inputs ($q'''_\mathrm{in}(t), \Delta p_\mathrm{pump}(t)$) are a function of time and will be defined by linear ramps that vary in ramp rate, maximum/minimum amplitude, and rest time between manipulations.
Additionally, there is one boundary condition,
\begin{equation}
    p(z=z_\mathrm{set},t) = p_\mathrm{loop}~,
\end{equation}
which is a static pressure boundary that defines the relative value of the pressure in the loop.

\begin{figure}
    \centering
    \begin{tabular}{cc}
       Control Input Permutations  & Exp. 8 Numerical Solution  \\
       \includegraphics[scale=0.7]{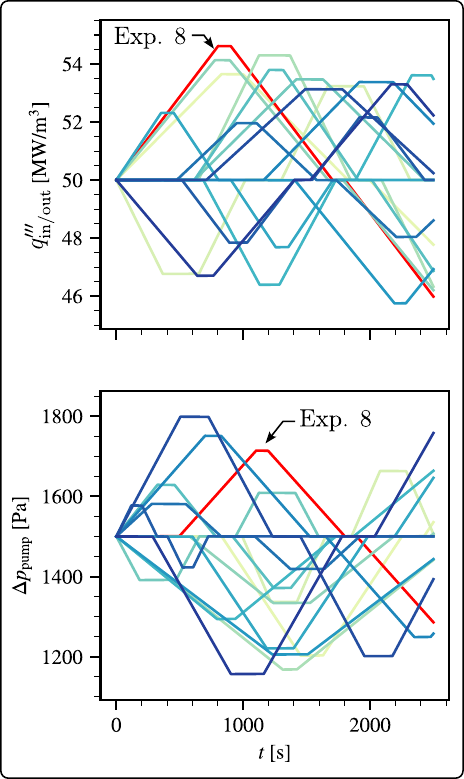} &
       \includegraphics[scale=0.7]{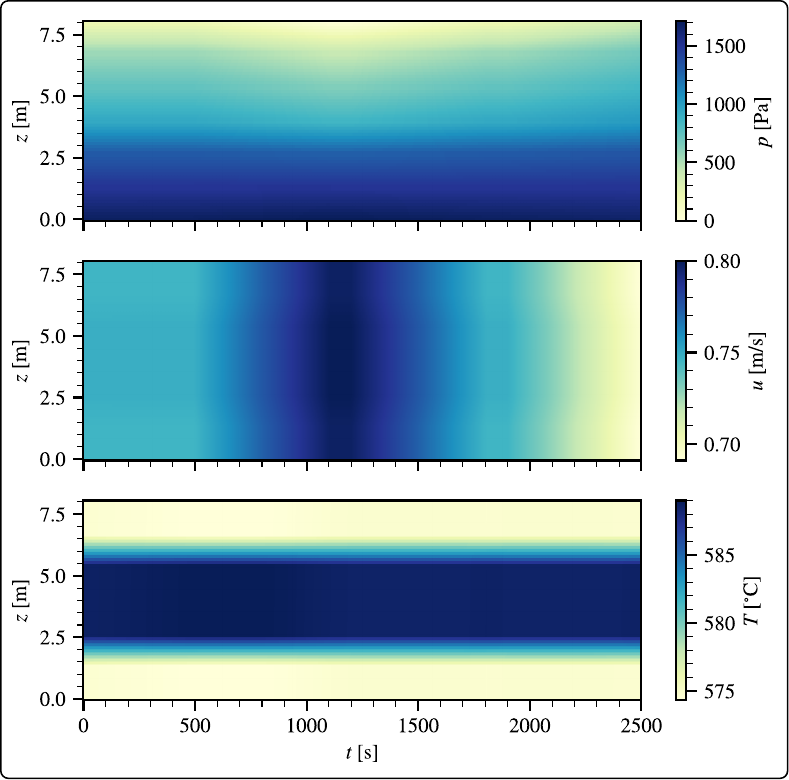}
    \end{tabular}
    \caption{An overview of the training dataset for the loop configuration.
    Left: Pump pressure jump and heat source inputs manipulated in time.
    The experiment chosen to display the solution is annotated ("Exp. 8").
    Right: Numerical solution of all field variables over the entire spatiotemporal domain.}
    \label{fig:exp_results_loop}
\end{figure}

Unlike the heated channel configuration, the loop configuration is more challenging from a dynamics modeling perspective.
First, there is no time-dependent manipulation of the field variables via boundary conditions. 
The manipulation of the pump head and heat source indirectly impact the field variables. 
Their evolution will have a spatiotemporal effect dictated by the transport equations, \cref{eq:mass,eq:mom,eq:nrg}. 
A visualization of the evolution of field variables is presented in \cref{fig:exp_results_loop}.
The pump head trajectory strongly impacts the velocity of the fluid.
Due to the control strategy, the temperature of the fluid remains approximately constant with respect to time but has a spatial variation. 

\subsection{Evaluation with test data}

\begin{figure}
    \centering
    \begin{tabular}{p{0.18\linewidth}p{0.4\linewidth}p{0.35\linewidth}}
         \centering Control Input & \centering Field values at fixed positions & \centering Field values at fixed times
    \end{tabular}
    \includegraphics[width=\linewidth]{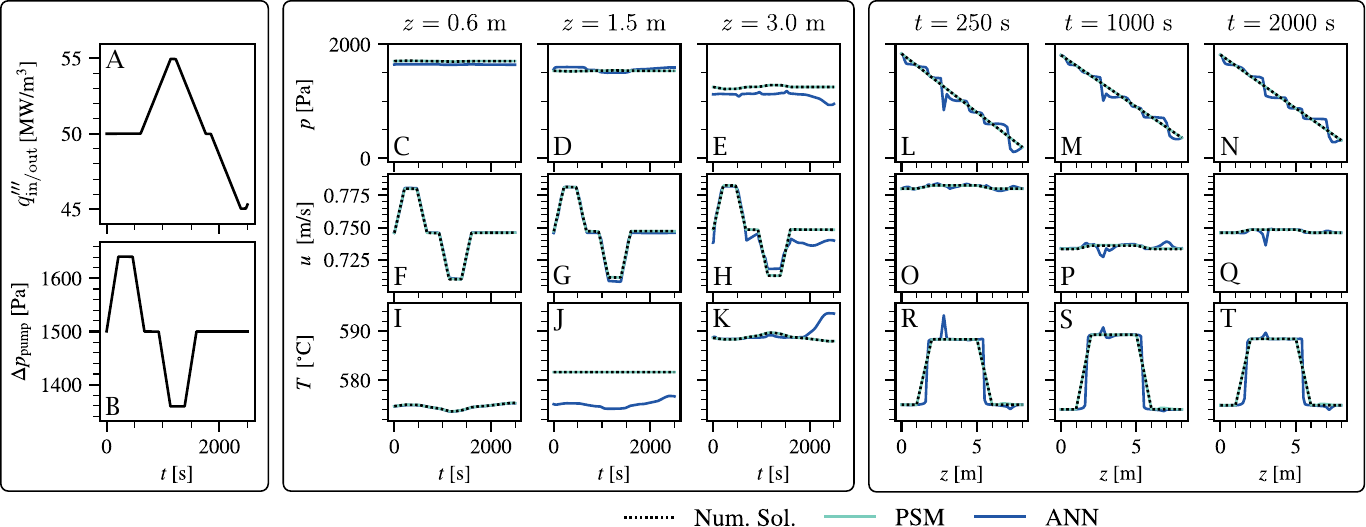}
    \caption{Contrasting the performance of PSM and ANN models in predicting the cooling system's evolution of pressure, velocity, and temperature.
        Subplots A-B: control input setting from the test dataset.
        Subplots C-K: field temporal evolution at fixed positions.
        Subplots L-T: field spatial distribution at fixed times.
    }
    \label{fig:results_loop}
\end{figure}

The performance of the PSM and ANN models for the cooling system is discussed next.
A sample of the control input permutations and the numerical solution of the field variables are presented in \cref{fig:results_loop}.
Through visual inspection, it is apparent that the ANN struggles with accurately predicting the qualitative evolution of all the field variables.
Because the ANN relies on a purely data-driven approach, the neural network correctly fits the field values about the spatial locations corresponding with sensors.
Yet, it fails to correctly interpolate the field values between the sensor locations -- because the problem is more complicated than the advection-dominated heated channel.
Whereas, the PSM has good quantitative and qualitative spatiotemporal prediction of all field variables.

\begin{figure}
\centering
\includegraphics[scale=0.78]{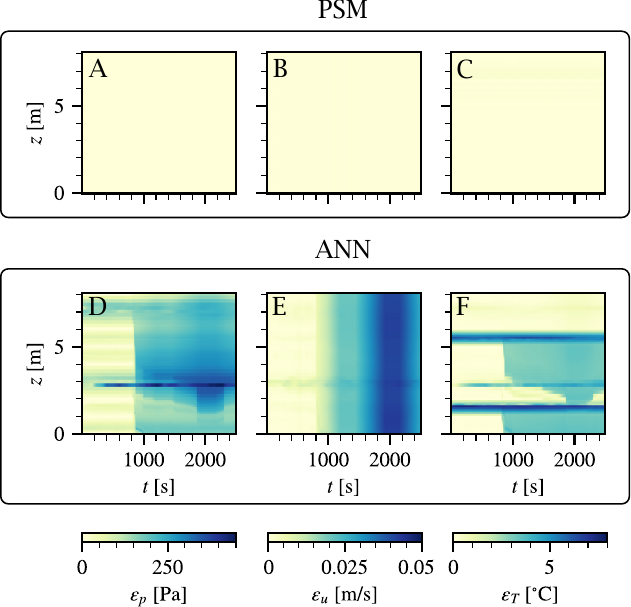}
\caption{Comparison of the RMSE in predicting the spatiotemporal evolution of all field variables, averaged over all test datasets.
Subplots A-C display error for the PSM model, and subplots D-F display error for the ANN model.
}
\label{fig:temp_se}
\end{figure}

The contrast between the ANN and PSM models is illustrated further in \cref{fig:temp_se}.
The figure presents error in predicting all field variables averaged over all test datasets.
The PSM model offers a significant performance advantage.
As the prediction horizon for the ANN increases, a significant error is introduced in all fields.
However, even for small time horizons, $t\leq\SI{100}{\second}$, there are significant errors in the prediction of temperature and pressure.
In summary, the results in this section show that in a complex experimental setting, a purely data-driven approach, \ie the ANN model, overfits to sensor measurements and results in unphysical predictions over the entire spatiotemporal domain.
In contrast, the PSM can generalize well to test data and achieve low error.

\FloatBarrier
\subsection{PSM Capability: Diagnostics}\label{subsec:results-diag}

Diagnostics in engineering systems are critical to maintaining operational efficiency and safety.
It involves the detection, identification, and characterization of faults or degradations that might occur. 
There are both model-free \citep{gertler2017fault} and model-based approaches \citep{isermann2005model} to diagnostics.
Model-based methods offer the advantage of leveraging the system's underlying physics and systematically interpreting deviations from expected behavior \citep{nguyen_physics-based_2022}.
Generally, in a model-based approach, we can identify the problem that has occurred based on a residual signature,
\begin{equation} \label{eq:residual}
    r = \mathcal{F}_\mathrm{nom}-\mathcal{F}_\mathrm{m}~,
\end{equation}
where $r$ is the computed residual value, and $\mathcal{F}_\mathrm{nom}$ and $\mathcal{F}_\mathrm{m}$ are the system model's nominal \vs measured values, respectively.
A non-zero valued $r$ indicates that the system's current performance is deviating from nominal.
For a system with multiple fields and components, $r$ would be multi-dimensional, generating a unique signature for each fault or degradation.
Suppose the model, $\mathcal{F}$, incorporates the fundamental physics of the system. 
In that case, this approach is powerful because it enables a physics-based interpretation of the degradation and enables us to discriminate between the faults that have occurred.

A key benefit of utilizing PSMs is their ability to incorporate the governing physics of the system through PDEs.
In a fully trained PSM, residuals from the mass, momentum, and energy PDEs, \cref{eq:mass,eq:mom,eq:nrg}, stabilize during the system's nominal fault-free operation.
If a component degrades during operation, these residuals will change and lead to a unique $r$ in \cref{eq:residual}.
An algorithm can then use these changes in the residual signature to identify and classify the nature of the degradation.

To validate this approach, a test case was designed to simulate a degradation in the cooling system's performance.
In this case, an abrupt partial blockage of the system occurs in Pipe 3 (immediately before the heat sink).
The blockage was simulated by increasing the friction factor, $f$ in \cref{eq:mom}, of that pipe by an order of magnitude.
In \cref{alg:deg}, a procedure is proposed to detect degradation and use a classification algorithm to diagnose the incident.
The algorithm requires continuous collection and analysis of new datasets, $\left(\mathcal{X}_\mathrm{new}, \mathcal{Y}_\mathrm{new}\right)$, and analyzing them.
It operates using two PSMs: a `nominal model', which represents the fault-free system, and a `twin model' whose parameters are updated once degradation is detected.
The update is triggered by a parameter, $\zeta$, in \cref{alg:deg:line:tau}.

After the twin model is updated, residuals from both PSMs are analyzed using a classification model to infer the type of degradation that has occurred.
The development of a classification model will be addressed in future work.

\begin{algorithm}
\caption{Degradation Detection and Classification}
\label{alg:deg}
\begin{algorithmic}[1]
\State \textbf{Input:} Original dataset $(\mathcal{X},\mathcal{Y})$, New dataset $(\mathcal{X}_\mathrm{new},\mathcal{Y}_\mathrm{new})$, Nominal PSM model $\mathcal{F}_\mathrm{nom}$, Threshold $\zeta$, Classification model $\mathcal{C}$
\State \textbf{Initialize:} Twin PSM model $\mathcal{F}_\mathrm{m}$, Degradation flag $\textit{latch} \gets \text{False}$
\ForAll{sample $(x_k, y_k)$ in $\left(\mathcal{X}_\mathrm{new},\mathcal{Y}_\mathrm{new}\right)$}
    \State Predict $\hat{y}_k = \mathcal{F}_\mathrm{nom}(x_k)$
    \State Compute $E = \frac{1}{n}\sum_{k=1}^{n}(\hat{y}_k - y_k)^2$
    \If{$E > \zeta$ \textbf{and} \textit{latch} is \textbf{False}}\label{alg:deg:line:tau}
        \State Set \textit{latch} to \textbf{True}
    \EndIf
    \If{\textit{latch} is \textbf{True}}
        \State Train $\mathcal{F}_{\text{m}}$ on $(x_k, y_k)$ without physics-informing
    \EndIf
\EndFor
\State Compute nominal residuals as $r_\mathrm{nom} = \text{PDEs}(\mathcal{F}_\mathrm{nom}(x))$ for $x$ in $\mathcal{X}$
\State Compute measured residuals as $r_\mathrm{m} = \text{PDEs}(\mathcal{F}_\mathrm{m}(x))$ for $x$ in $\mathcal{X}$
\State Calculate residual $r=r_\mathrm{nom}-r_\mathrm{m}$, in \cref{eq:residual}
\State Predict degradation type $d$ using $\mathcal{C}$ on $r$: $d = \mathcal{C}(r)$
\Return $d$
\end{algorithmic}
\end{algorithm}

\begin{figure}
    \centering
    \includegraphics[width=\linewidth]{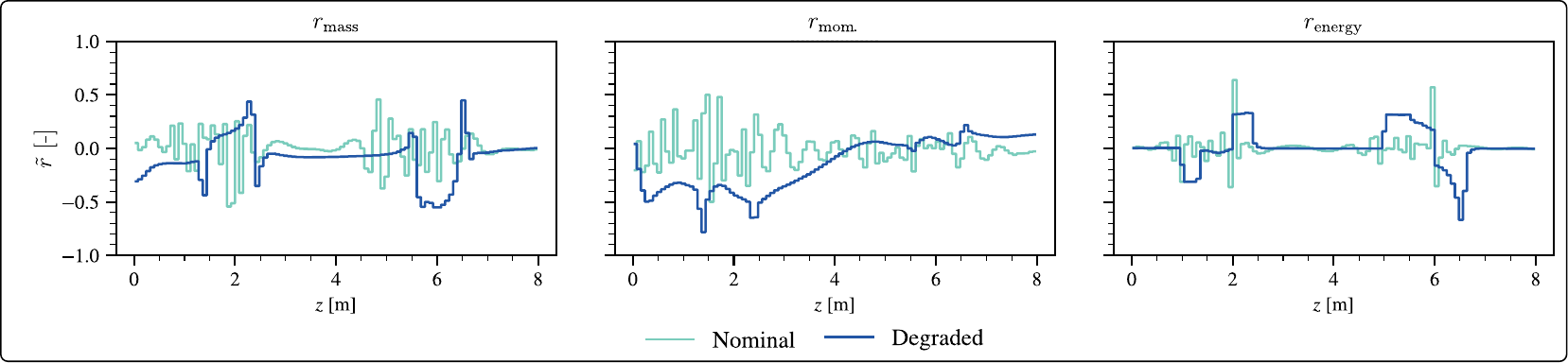}
    \caption{A comparison of the spatial distribution of nominal and degraded residuals, $r$ and $r_\mathrm{deg}$, respectively, in \cref{alg:deg}. The residual values have been scaled by their maximum and minimum value. }
    \label{fig:deg}
\end{figure}

The application of \cref{alg:deg} to the data generated from the degradation test case is discussed next.
In \cref{fig:deg}, the scaled residual values are contrasted for nominal and degraded models.
The residuals are presented as a function of position.
Within the cooling system, Pipe 3 is located at $\SI{4.0}{\meter}\leq z \leq \SI{5.0}{\meter}$.
There are a few important observations to consider.
First, the nominal model has a relatively constant spatial distribution for the momentum residual, $r_\mathrm{mom.}$.
Whereas, for the degraded model, the spatial distribution shows a qualitative shift in the residual around the location of Pipe 3. 
Second, the mass and energy residuals do not show a significant shift in the spatial distribution of residuals.
From these observations, a classification algorithm can be trained to identify the degradation that has occurred.
In summary, in this section, an algorithm was proposed and utilized on a test case to show that PSMs can provide meaningful data to a diagnostics algorithm to identify system degradation or fault.

\section{Discussion and Concluding Remarks} \label{sec:disc}

Autonomous systems are envisioned to achieve real-time optimization, be flexible in operation, and be fault tolerant.
These capabilities necessitate a model-based approach.
However, purely data-driven methods lack physical constraints like mass conservation.
Consequently, model-free approaches, such as PID control, dominate the operation of modern transport-dominated systems such as chemical, biomedical, and power plants.
To address this challenge, we propose the adoption of PSMs.

A PSM is a deep neural network trained by fusing sensor data with physics-informing using the components’ PDEs.
The result is a physics-constrained end-to-end differentiable forward dynamics model.
Two separate experiments were designed, in silico, to demonstrate PSM capabilities for transport-dominated systems.
The first experiment was a heated channel, and the second was a cooling system loop.
The datasets used to train the PSM models are available online\footnote{Dataset will be released on journal acceptance.}.
Evaluation on test data indicates that the PSM approach is quantitatively and qualitatively more accurate than a purely data-driven ANN.
Additionally, both homo- and heteroscedastic noise were introduced during training to demonstrate the applicability of PSMs to a physical plant.

Owing to their design as end-to-end differentiable state-space models, PSMs demonstrated a multitask capability:
\begin{enumerate}
    \item 
    The first demonstration used a PSM model to construct a supervisory controller (\cref{subsec:results-control}).
    To achieve this, the proposed NCG algorithm, \cref{alg:NCG}, used a linear state-space representation sequentially updated through Jacobian Linearization of the PSM.
    The capability was utilized to demonstrate both constant and time-dependent constraint enforcement.
    Thus, PSMs can be utilized for control schemes that require a nonlinear or linear state-space representation.
    \item 
    The second demonstration proposed a diagnostics algorithm that utilized the PSM as a basis (\cref{subsec:results-diag}).
    Conservation laws, in the form of PDEs, can be evaluated using the PSM.
    The proposed algorithm, \cref{alg:deg}, uses residuals from each of the PDEs, which are available as a function of location. 
    The demonstration simulated a partial blockage of a pipe in the cooling system.
    Two separate PSMs were used to obtain PDE residuals, one trained on the nominal dataset and one that had been transfer-learned on the degraded dataset.
    It was demonstrated that the differences in the residuals of the momentum equation can then be used as inputs to a classification algorithm.
    \item
    An additional use-case of PSMs, not covered in this work, is as a basis for Digital Twins (DTs) \citep{jones2020characterising}.
    DTs require that the ``digital'' model used to represent the physical asset is updated frequently through feedback from measurements to match drifts from anticipated dynamics \eg due to wear and tear.
    Because the PSM is end-to-end differentiable, this feature can be accommodated through online learning of PDE coefficients.
\end{enumerate}

There are potential drawbacks of the PSM that warrant further research.
First, this work did not rigorously study the number and placement of sensors (measurement locations).
The optimal choice would be highly dependent on the characteristics of the system.
If a PSM is applied to an existing facility, adding additional sensors may not be possible.
Second, the current work intentionally focused on relatively small systems with few components.
For large systems, \eg large power plants with multiple loops, a single PSM would not be sufficient to model the entire facility.
Additional methods are needed to train and coordinate multiple PSMs representing separate parts of a plant.
A potential approach may be to adopt a graphical architecture where PSMs represent transport between nodes.

\nomenclature[A]{PSM}{Physics-informed State-space neural network Model}
\nomenclature[A]{PDE}{Partial Differential Equation}
\nomenclature[A]{ANN}{Artificial Neural Network}
\nomenclature[A]{PINN}{Physics-Informed Neural Network}
\nomenclature[A]{NCG}{Neural Command Governor}
\nomenclature[A]{RG}{Reference Governor}
\nomenclature[A]{BC}{Boundary Condition}
\nomenclature[A]{CG}{Command Governor}
\nomenclature[A]{PID}{Proportional-Integral-Derivative controller}
\nomenclature[A]{MSE}{Mean Squared Error}
\nomenclature[A]{RMSE}{Root-Mean-Square Error}

\nomenclature[G]{$\sigma$}{Standard deviation of homoscedastic noise, \cref{eq:homo-noise}}
\nomenclature[G]{$\xi$}{Variance factor of heteroscedastic noise, \cref{eq:hetero-noise}}
\nomenclature[G]{$\rho$}{Fluid density, \cref{eq:mass,eq:mom,eq:nrg}}
\nomenclature[G]{$\theta$}{Set of parameters in a neural network, \cref{eq:nn_def}}
\nomenclature[G]{$\Omega$}{Space domain, \cref{eq:nn_def}}
\nomenclature[G]{$\tau$}{Time domain, \cref{eq:nn_def}}
\nomenclature[G]{$\mathcal{L}_\Sigma$}{Total loss, \cref{eq:loss_psm}}
\nomenclature[G]{$\alpha, \beta$}{Weighting coefficients for total loss, \cref{eq:loss_psm}}

\nomenclature[R]{$\mathcal{L}_m$}{Measurement loss, \cref{eq:meas_loss}}
\nomenclature[R]{$\mathcal{L}_p$}{Physics-informed loss, \cref{eq:pinn_loss}}
\nomenclature[R]{$\mathcal{F}$}{Neural-network model (ANN, or PSM)}
\nomenclature[R]{$\mathcal{X}_m$}{Measurement input dataset, \cref{eq:ds1}}
\nomenclature[R]{$\mathcal{X}_p$}{Physics-informed input dataset, \cref{eq:ds_pinn}}
\nomenclature[R]{$A,B$}{Linear state-space matrices, \cref{eq:state,eq:jacA,eq:jaqB}}
\nomenclature[R]{$O_\infty$}{Maximal output admissible set, \cref{eq:moas}}
\nomenclature[R]{$x$}{State}
\nomenclature[R]{$v$}{Control input}
\nomenclature[R]{$z$}{Position}
\nomenclature[R]{$t$}{Time}
\nomenclature[R]{$u$}{Fluid speed, \cref{eq:mom}}
\nomenclature[R]{$f$}{Friction factor, \cref{eq:mom}}
\nomenclature[R]{$T$}{Fluid temperature, \cref{eq:nrg}}
\nomenclature[R]{$q'''$}{Volumetric heat source, \cref{eq:nrg}}
\nomenclature[R]{$p$}{Fluid pressure, \cref{eq:mom}}
\nomenclature[R]{$r_k$}{RG reference input at step $k$, \cref{eq:rg1}}
\nomenclature[R]{$x_{00}$, $v_{00}$}{State and input values at the linearization point, \cref{eq:lpv,eq:lpx}}
\nomenclature[R]{$r$}{Residuals during diagnostics, \cref{eq:residual}}
\nomenclature[R]{$f_{\mathrm{loss}}$}{Loss function used in network training, \cref{eq:meas_loss,eq:pinn_loss,eq:lcl}}

\printnomenclature

\section*{CRediT authorship contribution statement}
\textbf{Akshay J.~Dave}: Conceptualization, Methodology, Writing - Original Draft, Funding acquisition.
\textbf{Richard B.~Vilim:} Supervision, Methodology, Writing - Review \& Editing.

\section*{Acknowledgements}
Argonne National Laboratory supports this work under the Laboratory Directed Research \& Development Program, Project 2022-0077.

\bibliographystyle{unsrtnat}
\bibliography{references}

\end{document}